\documentclass[journal,twoside]{IEEEtran}

\usepackage{url}
\usepackage{cite}
\usepackage{array}
\usepackage{color}
\usepackage{xcolor}
\usepackage{colortbl}
\usepackage{hyperref}
\usepackage{multirow}
\usepackage{makecell}
\usepackage{booktabs}
\usepackage{siunitx}
\usepackage{algorithm}
\usepackage{algorithmic}
\usepackage[]{amssymb}
\usepackage[cmex10]{amsmath}
\usepackage{float}

\ifCLASSINFOpdf
  \usepackage[pdftex]{graphicx}
\else
  \usepackage[dvips]{graphicx}
\fi
\ifCLASSOPTIONcompsoc
\usepackage[caption=false,font=normalsize,labelfont=sf,textfont=sf]{subfig}
\else
 \usepackage[caption=false,font=footnotesize]{subfig}
\fi

\renewcommand{\thesubsection}{\Alph{subsection}}
\renewcommand\thetable{\Roman{table}}

\graphicspath{{./figures/}{}}

\DeclareGraphicsExtensions{.pdf, .png}

\usepackage{soul}

\hypersetup{
    colorlinks=true,
    linkcolor=red,
    citecolor=green,
    urlcolor=magenta,
}
\begin{document}

\title{S\textsuperscript{4}DL: Shift-sensitive Spatial-Spectral Disentangling Learning for Hyperspectral Image Unsupervised Domain Adaptation}

\author{
Jie~Feng,~\IEEEmembership{Senior~Member,~IEEE,}
Tianshu~Zhang, Junpeng Zhang, 
Ronghua~Shang, Weisheng Dong,~\IEEEmembership{Senior~Member,~IEEE,} Guangming Shi,~\IEEEmembership{Fellow,~IEEE}, and~Licheng~Jiao,~\IEEEmembership{Fellow,~IEEE} 
        
\thanks{This work was supported in part by the National Natural Science Foundation of China under Grant 62271374, Grant 62176200, Grant 62077038, and Grant 62176196; in part by the State Key Program of National Natural Science of China under Grant 61836009; in part by the Natural Science Basic Research Program of Shaanxi under Grant 2022GY-065 and Grant 2022JC-45; in part by the Fundamental Research Funds for the Central Universities under Grant QTZX23047.}

\thanks{J. Feng, T. Zhang, J. Zhang, R. Shang, W. Dong, G. Shi and L. Jiao are with the Key Laboratory of Intelligent Perception and Image Understanding of Ministry of Education of China, Xidian University, Xi'an 710071, P.R. China (e-mail: jiefeng0109@163.com; zts.xidian@gmail.com; junpengzhang@xidian.edu.cn;rhshang@mail.xidian.edu.cn; wsdong@mail.xidian.edu.cn;gmshi@xidian.edu.cn; lchjiao@mail.xidian.edu.cn).}}

\markboth{IEEE TRANSACTIONS ON NEURAL NETWORKS AND LEARNING SYSTEMS, VOL. XX, NO. X, April 2024}%
{Feng \MakeLowercase{\textit{et al.}}: }

\maketitle
\begin{abstract}
Unsupervised domain adaptation techniques, extensively studied in hyperspectral image (HSI) classification, aim to use labeled source domain data and unlabeled target domain data to learn domain invariant features for cross-scene classification. 
Compared to natural images,  numerous spectral bands of HSIs provide abundant semantic information, but they also increase the domain shift significantly. 
In most existing methods, both explicit alignment and implicit alignment simply align feature distribution, ignoring domain information in the spectrum.
We noted that when the spectral channel between source and target domains is distinguished obviously, the transfer performance of these methods tends to deteriorate.
Additionally, their performance fluctuates greatly owing to the varying domain shifts across various datasets. To address these problems, a novel shift-sensitive spatial-spectral disentangling learning (S\textsuperscript{4}DL) approach is proposed. 
In S\textsuperscript{4}DL,  gradient-guided spatial-spectral decomposition is designed to separate domain-specific and domain-invariant representations by generating tailored masks under the guidance of the gradient from domain classification. 
A shift-sensitive adaptive monitor is defined to adjust the intensity of disentangling according to the magnitude of domain shift. 
Furthermore, a reversible neural network is constructed to retain domain information that lies in not only in semantic but also the shallow-level detailed information. 
Extensive experimental results on several cross-scene HSI datasets consistently verified that S\textsuperscript{4}DL is better than the state-of-the-art UDA methods. Our source code will be available at \url{https://github.com/xdu-jjgs/S4DL}.

\end{abstract}


\ifCLASSOPTIONpeerreview
\begin{center} \bfseries EDICS Category: 3-BBND \end{center}
\fi
\IEEEpeerreviewmaketitle

\section{Introduction}
\label{sec:intro}

\IEEEPARstart{H}{yperspectral} image (HSI) is obtained by capturing information from the reflection  of light from an object or scene at hundreds of different wavelengths. Unlike widely-adopted RGB images, each pixel in HSI not only contains visible light information, but also covers near-infrared, short-wave infrared, mid-infrared and long-wave infrared information, which enables HSIs to capture richer spectral information and detect more accurately~\cite{9463743}. 

However, factors like lighting, seasonal variations, atmospheric conditions, and differences in sensors lead to an inevitable domain shift between HSI of different scenes, undermining the assumption of independent and identically distributed data. 
This limitation hinders the transferability and generalization of traditional classification models to other scenes. 
In response to these challenges, unsupervised domain adaptation (UDA) for HSI has been introduced~\cite{zhang2021topological, li2023supervised, huang2023cross}, which seeks to apply knowledge from a labeled source domain to an unlabeled target domain. 
While the target task and label spaces of the training and test sets remain identical, their feature distributions differ yet are related. 
Therefore, the model needs to learn domain-invariant features while dealing with the target task to achieve cross-scene HSI classification.

UDA is one of the most effective solutions for cross-scene HSI classification by extracting domain-invariant features.
Inspired by disentangling learning, domain disentangling methods explicitly separate domain-invariant feature and domain-specific feature while maintaining the model transferability by seeking for the alignment on the domain-invariant features between the source and target domains~\cite{bousmalis2016domain, zhou2023self, wang2023generalized}. Based on this framework, existing methods introduce style information~\cite{lee2021dranet} and instance information~\cite{liu2022decompose} in disentangling stage to facilitate the feature disentanglement. Nevertheless, for handling cross-scene HSI classification, it is necessary to capitalize on the domain information in a large number of spectral bands embedded in HSIs.

\begin{figure}[ht]
    \centering
    \includegraphics[width=\linewidth]{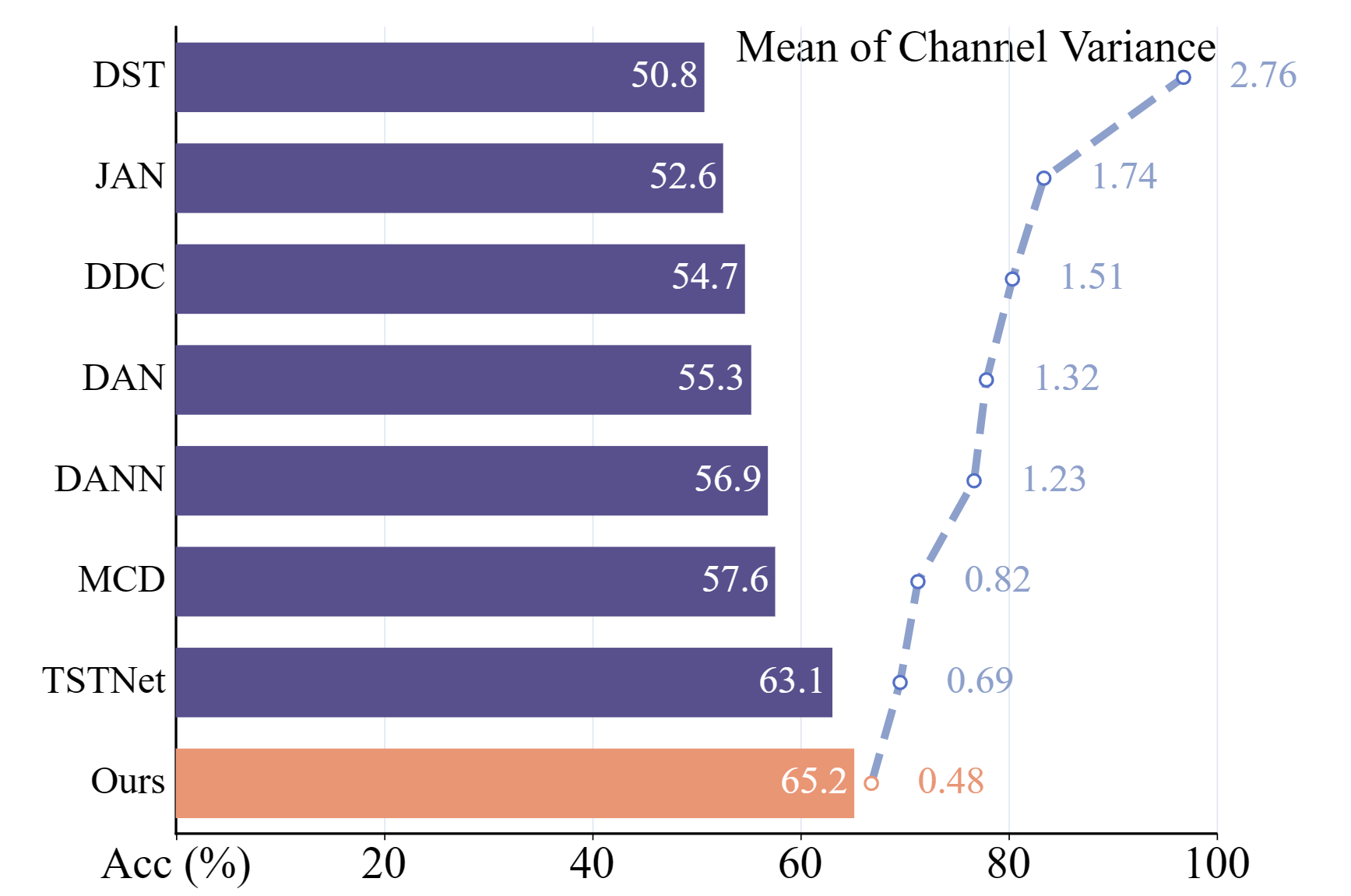}
    \caption{Channel Variance and Model Performance. 
    It shows mean of standard deviations of spectral channels and corresponding classification accuracies of UDA methods on the HyRANK dataset, where the standard deviation was computed by the activation values of feature maps for source and target domain data. 
    The bar graph delineates accuracy, while the line graph reflects the mean inter-domain standard deviation of the model's channels. The experiments were conducted on the HyRANK dataset~\cite{Karantzalos2018}.}
    \label{fig:accwithstd}
\end{figure}

In HSI, there may be a significant domain gap between domains. This is due to the spectral uncertainties of targets, which can be remarkably significant due to the complexity of spectral information and variations in the imaging environment. As a result, the extracted features can become confused across different spectral channels. This channel confusion makes it difficult to consistently extract invariant features, ultimately affecting the transferability of the model across domains. Thus, the domain gap induced by spectral variations undermines the stability and effectiveness of domain-invariant feature extraction.
The variation and accuracy of this phenomenon have been scrutinized through different methods, in Fig.~\ref{fig:accwithstd}, the differences in spectrum could cause the model to behave inconsistently between the source domain and the target domain, increasing the standard deviation of the activation values between domains, consequently leading to inferior transfer results. Therefore, it is crucial to reduce domain shifts existing in the spectral dimension of HSIs.

Furthermore, domain shifts in HSIs are caused by various factors including imaging time, imaging geographical location, imaging season, etc.~\cite{tang2022unsupervised, zhang2023multi}, therefore the degree of domain shifts is diverse in different scenes and different training stages. Traditional fixed alignment strategy across different datasets and training phases may result in insufficient transfer or negative transfer in HSIs.

In this paper, we propose a novel framework, named shift-sensitive spatial-spectral disentangling learning network (S\textsuperscript{4}DL), which aims to enhance the disentanglement of domain-invariant features from domain-specific features. 
Firstly, a gradient-guided spatial-spectral decomposition is designed to quantify the domain information of each channel based on the gradient of domain classification, and subsequently generate tailored masks to decouple domain-invariant and domain-specific channels.
Secondly, a shift-sensitive adaptive monitor is incorporated to cope with various degrees of domain shift in various datasets and training stages. This detector continually monitors the inter-domain variance  and dynamically fine-tunes the alignment strategy by using exponential moving average (EMA) strategy.
Finally, a reversible feature extractor (RFE) is constructed to retain domain information lying in low-level features by preserving and embedding low-level features together with semantic features for alignment. 

Our contributions are summarized as follows:
\begin{enumerate}
    \item We propose a novel joint disentangling unsupervised domain adaptation framework for cross-scene HSI classification by collaboratively decoupling both spectral and spatial dimensions simultaneously, and RFE is introduced for enhancing the fine-grained spatial information at high-level feature maps, which jointly leads to improved transferability to different scenes.

    \item In feature disentangling, the gradient-based calculation associated with domain classifier provides a direct measure of domain-specific information for each channel, allowing for continuous monitoring and dynamic disentangling domain-invariant channels for classification. 

    \item To address the variations in the extent of domain gap across different scenes and different training phases, we propose an adaptive domain shift detector that dynamically modifies the model's alignment strategy according to the scale of domain shifts during training, enabling it to be suitable for different datasets with various domain shifts, thus enhancing the model's generalization capabilities.
    
\end{enumerate}

\section{Related Works}

\subsection{Unsupervised Domain Adaptation}
The primary goal of UDA is to leverage the extensive knowledge gained from a source domain, characterized by abundant annotated training examples, for application in target domains that exclusively utilize unlabeled data.
To achieve such a goal, a variety of methods are built by matching the statistical distribution differences~\cite{tzeng2014deep, long2015learning, long2017deep, sun2016deep, zellinger2017central, kang2019contrastive}, aligning marginal or joint distribution~\cite{ganin2015unsupervised, ganin2016domain, long2018conditional, saito2018maximum, zhang2019bridging} or adopting self-training schemes~\cite{zou2018unsupervised, zou2019confidence, liu2021cycle} and consistency regularizations~\cite{french2017self, tarvainen2017mean, ge2020mutual}.

\begin{figure*}[ht]
    \centering
    \includegraphics[width=0.9\linewidth]{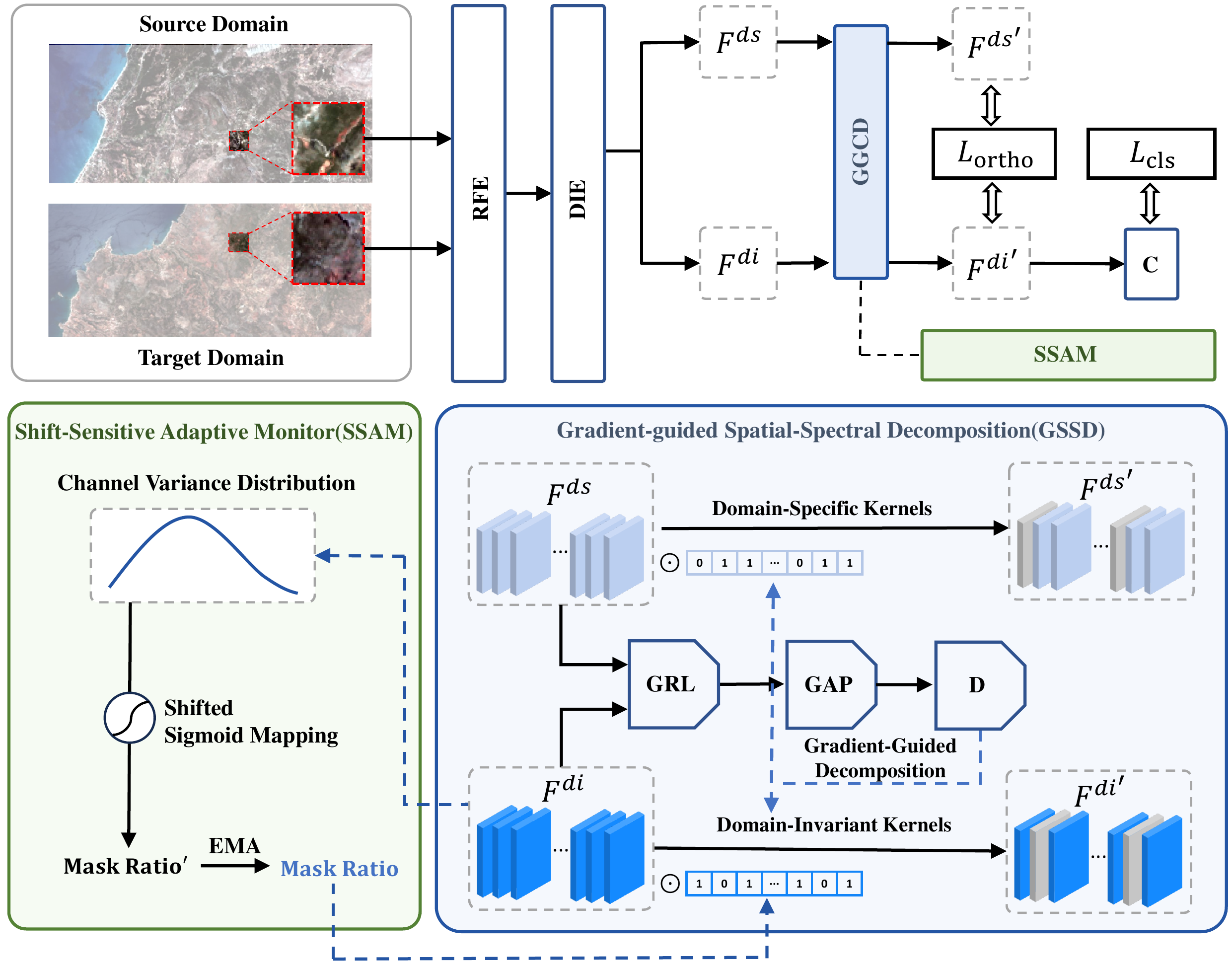}
    \caption{Framework of the proposed S\textsuperscript{4}DL, including  Reversible Feature Extractor, Gradient-guided Spatial-Spectral Decomposition and Shift-Sensitive Adaptive Monitor.}
    \label{fig:model-overall}
\end{figure*}

Recently, domain disentangling defines a new UDA paradigm by separating  domain-invariant and domain-specific features into distinct feature subspaces, while maintaining the orthogonality between the shared subspace for domain-invariant information and the private subspace for the domain-specific information~\cite{bousmalis2016domain, wu2021vector, wang2023generalized}. 
This design promotes the transfer of domain-invariant features to downstream tasks and discards the harmful domain-specific features, which guarantees the model’s ability to transfer and generalize by performing domain alignment in the shared subspace. 
Specifically, Bousmalis et al.~\cite{bousmalis2016domain} proposed Domain Separation Networks (DSN), firstly introduced disentangled representation learning to domain adaptation, extracting source-specific features, target specific features and domain-invariant features by private source encoder, private target encoder, and shared encoder, respectively. 
Then, a shared decoder is used to reconstruct the original images. These three encoders are decoupled by the orthogonal loss. 
Building on DSN, Lee et al.~\cite{lee2021dranet} attempted to
disentangle individual features by content and style, then translate domains by style transformations.
In order to narrow intra-domain and inter-domain gaps, Zhou et al.~\cite{zhou2023self} proposed self-adversarial disentangling to learn domain-invariant features in a domain-specific dimension.
However, these disentangling strategies have not paid attention to the domain information in channels. Nevertheless, there is plenty of spectral information in HSIs, which causes the insufficiency of existing disentangling methods, leading to a performance decline in cross-scene classification of HSIs.

\subsection{Domain Adaptation for Hyperspectral Image Classifcation}

In domain adaptation of HSI classification, previous works focus on learning more discriminative features in an unsupervised way, especially in the target domain. For example, Zhang et al.~\cite{zhang2021topological} proposed topological structure and semantic information transfer network (TSTNet) to capture complex topological relationships in HSIs. It models the topological relationships among HSIs as a graph optimal transmission problem, introducing a graph convolutional network (GCN) and a convolutional neural network (CNN) to perform joint classification, constraining their outputs with consistency loss. Ning et al.~\cite{ning2023contrastive} introduced class alignment in the contrastive learning framework, which helps in learning instance-level discriminative features for classification. Further, in order to reduce the negative effect of abnormal samples on the learning of discriminatory features, Ning et al.~\cite{ning2024domain} proposed compact prototype contrast adaptation, alleviating the negative impact of outliers. Similarly, Li et al~\cite{li2023supervised} introduced a supervised contrastive learning-based approach to enhance the separability of the intra-domain HSI data. However, different from these methods, our method takes feature disentangling as a starting point, learning how to explicitly separate discriminative features from other features, to avoid being negatively impacted by domain shift information during transfer.

\subsection{Channel-wise Feature Enhancement}
Work that explicitly models the importance weights of channels is also relevant to this paper, for example, the squeeze-and-excitation mechanism~\cite{hu2018squeeze} and the channel attention mechanism~\cite{woo2018cbam}. These methods have been applied across different domains, including semantic segmentation~\cite{fu2019dual} and image super-resolution~\cite{zhang2018image}. 
Contrary to these aforementioned methods that designed singularly for feature extraction within a specific domain, our method diverges in two aspects. Firstly, in terms of channel importance generation, instead of the SE module or an attention matrix, we quantify the domain information of each channel explicitly through the gradient of the domain classification. Secondly, in terms of utilization, as opposed to their emphasized on enhancing feature extraction within a singular domain, we employ it for feature disentanglement. This involves the explicit decomposition of domain-invariant and domain-specific channels, which aim to amplify the inter-domain transfer ability of the model.

\section{Methodology}

\textcolor{black}{To handle the insufficient disentangling and the stationary alignment strategies in the existing methods, we propose a novel shift-sensitive spatial-spectral disentangling learning network, namely S\textsuperscript{4}DL.}
Our model comprises three main components: the reversible feature extractor (RFE), the gradient-guided spatial-spectral decomposition (GSSD), and the shift-sensitive adaptive detector (SSAM). 

As illustrated in Fig.~\ref{fig:model-overall}, 
\textcolor{black}{our S\textsuperscript{4}DL deploys a siamese architecture for feature extraction.
For a given pair of images from source and target domains, their corresponding feature maps are extracted using a shared backbone.
For preventing the vanishing of low-level information at high-level features, we substitute the conventional CNN backbone with the RFE.
}

\textcolor{black}{The obtained feature maps are then fed to the domain-invariant extractor, and each feature map $\boldsymbol{F}$ is disentangled into a domain-invariant feature map $\boldsymbol{F}^{di}$ and its supplementary domain-specific counterpart $\boldsymbol{F}^{ds}$, such that $\boldsymbol{F} = \boldsymbol{F}^{di} + \boldsymbol{F}^{ds}$.
For quantifying the domain information across diverse channels, the proposed GSSD is attached to $(\boldsymbol{F}^{di}, \boldsymbol{F}^{ds})$ for further refining the obtained domain-invariant and domain-specific components along the spectral dimension, with the assistance of the gradient back-propagated from a domain discriminator $\boldsymbol{D}$. 
For enhancing the adaptivity of our GSSD to fluctuative domain shifts across scenes, the proposed SSAM is injected for dynamically adjusting the proposed GSSD.
Finally, the obtained domain-invariant feature is fed to the classification head.
}

\subsection{Gradient-guided Spatial-Spectral Decomposition}

With extended spectrum coverage and \textcolor{black}{dense spectral sampling interval}, HSIs 
provide rich channel dimensional information, compared to natural images.  While existing UDA methods are dominantly constructed on an over-simplified encoder for extracting domain-invariant features and their domain-specific counterpart, the underlying structure and distribution along the channel dimension are overlooked. 
As summarized in Fig. \ref{fig:accwithstd}, it is evident that, \textcolor{black}{without proper treatments for handling the channel information}, considerable variances are observed over the obtained domain-invariant features, which hinders model transferability across domains.
To this end, we highlight that a stronger channel disentangling mechanism is key to extracting domain-independent features for hyperspectral image domain adaptation.
In this work, we propose a novel GSSD module, where refinements along the channel dimension are attended to the decoupled domain-specific and domain-invariant features. 
Since it is non-trivial to conduct such refinements with no explicit supervision available, we dive into the gradients from a domain classifier for additional guidance, leading to improved domain-invariant features with minimized channel variance.

Specially, for an input image from either the source domain or the target domain, let $\mathbf{F} \in \mathbb{R}^{H \times W \times C}$ denote its corresponding feature map from the backbone, where $H$ and $W$ are its height and width, and $C$ is the number of channels.
This feature map is fed to the domain invariant encoder (DIE) and decomposed to a domain-invariant component and its domain-specific counterpart, denoted by $\mathbf{F}^{di}$ and $\mathbf{F}^{ds}$, respectively. Notably, $\mathbf{F}^{ds}=\mathbf{F}-\mathbf{F}^{di}$.
Instead of directly passing $\mathbf{F}^{di}$ and $\mathbf{F}^{ds}$ for down-stream tasks, we introduce two $C$-dimensional binary channel filters, termed as the domain-invariant kernels $\mathbf{u} \in \mathbb{R}^{C}$ and the domain-specific kernels $\mathbf{v} \in \mathbb{R}^{C}$.
Then, the refined domain-invariant feature $\tilde{\mathbf{F}}^{di}$ and domain-specific feature $\tilde{\mathbf{F}}^{ds}$ can be obtained by attending these filters to each pixel location at $\mathbf{F}^{di}$ and $\mathbf{F}^{ds}$, respectively. 
This process can be achieved by applying $1 \times 1$ depth-wise convolution filters over $\mathbf{F}^{di}$ and $\mathbf{F}^{ds}$ with the kernels constructed from $\mathbf{u}$ and $\mathbf{v}$,
\begin{equation}
\label{eq:dwconv-u}
\begin{aligned}
\tilde{\mathbf{F}}^{di} = \text{DWConv}_{1 \times 1}(\mathbf{F}^{di}, \mathbf{u}), \\
\tilde{\mathbf{F}}^{ds} = \text{DWConv}_{1 \times 1}(\mathbf{F}^{ds}, \mathbf{v}),
\end{aligned}
\end{equation}
where $\text{DWConv}_{1 \times 1}(\cdot, \cdot)$ refers to the depth-wise convolution operator with a kernel of size of $1 \times 1$.
Intuitively, these kernels play important roles for filtering out non-disentanglable channels, however, the selection of proper kernels remains an open problem, due to the inaccessibility of supervision on both kernels.

As each channel contributes to feature in-variance by varying significance, we prefer to gain guidance from this fact for quantifying the \textcolor{black}{amount} of domain information in each channel.
To this end, a domain discriminator $D$ is employed for determining the domain labels of the domain-invariant features from source and target domains, and our domain-invariant filter is estimated on the contribution of each channel toward accurate domain prediction.
For a given domain-invariant feature map $\mathbf{F}^{di}$ obtained by DIE, it is globally pooled into a feature vector $\mathbf{P}^{di} \in \mathbb{R}^{C}$, which is then passed to the domain discriminator $D$.
For measuring the contribution of a feature channel toward accurate domain prediction, we define a \textcolor{black}{channel domain discriminability metric of \(\mathbf{F}^{di}\)} as
\begin{equation}
\mathbf{w}_{c}^{di} = 
\mathbf{P}^{di}_{c}
\cdot 
\frac{\partial D(\mathbf{P}^{di})}{\partial \mathbf{P}^{di}_c},
\forall c \in \{1, 2, 3, \ldots, C\},
\end{equation}
where a greater $\mathbf{w}_{c}^{di}$ naturally implies the $c$-th channel contains more domain-discriminative information.
Our domain-invariant filter $\mathbf{u}$ is designed for suppressing those channels with top \textcolor{black}{$K$}  \(\mathbf{w}^{di}\) scores as
\begin{equation}
\begin{aligned}
\mathbf{u}_c &= \begin{cases}
0 & \textrm{if } c \in \textrm{argsort}(-\mathbf{w}^{di})[:K] \textrm{ and } \mathbf{w}^{di}_c > 0 \\
1 & \textrm{otherwise}
\end{cases},
\end{aligned}
\end{equation}
where the kernel elements of a top portion of channels are set as 0.
 \(\textrm{argsort}(\cdot)[:K]\) is used to find the indices of the smallest $K$ elements.
The number of suppressed channels $K$ is a fraction of the total channel number, $K = C \times r$, where $r$ denotes the suppressing ratio.
This design helps ease the domain-invariant feature refinement among channel dimension by filtering out channels with strong contributions to domain-specific information.

At the same time, as each channel contributes to specific features by varying contributions, we use the shared domain discriminator \(D\) to provide guidance for quantifying the  \textcolor{black}{amount} of domain information in each channel. Based on the gradient of each channel obtained from different domain labels, the contribution of each channel to domain classification is estimated, thereby our domain-specific filter \(\mathbf{v}\) is generated. The domain-specific feature map \(\mathbf{F}^{ds}\) extracted by DIE  is processed through GAP, after which $\mathbf{P}^{ds} \in \mathbb{R}^{C}$ is input into \(D\). \textcolor{black}{The channel domain discriminability metric of $\mathbf{F}^{ds}$}  is defined correspondingly to measure the contribution of each channel in \(\mathbf{F}^{ds}\) towards accurate domain classification. 

\begin{equation}
\mathbf{w}_{c}^{ds} = 
\mathbf{P}^{ds}_{c}
\cdot 
\frac{\partial D(\mathbf{P}^{ds})}{\partial \mathbf{P}^{ds}_c},
\forall c \in \{1, 2, 3, \ldots, C\},
\end{equation}
where a greater $\mathbf{w}_{c}^{ds}$  implies the $c$-th channel in \(\mathbf{F}^{ds}\) contains more domain-discriminative information in the same way. Conversely, the domain-specific filter $\mathbf{v}$ is designed for suppressing those channels with the smallest \textcolor{black}{$K$}  absolute scores of \(\mathbf{w}^{ds}\) as
\begin{equation}
\begin{aligned}
    \mathbf{v}_c &= 
    \begin{cases}
    0 & \text{if } c \in \text{argsort}(\lvert \mathbf{w}^{ds} \rvert)[:K], \\
    1 & \text{otherwise}
    \end{cases}
\end{aligned},
\end{equation}
where \(|\cdot|\) represents the absolute value operation to avoid confusing incorrect domain-specific information with domain-invariant information, and the kernel elements of the smallest  \textcolor{black}{$K$}  absolute values  of channels are set to 0. This further decomposition of \(\mathbf{F}^{ds}\) suppresses domain-invariant channels and preserves domain-specific channels, which helps remove domain-specific feature along the channel dimension by identifying channels with little contribution to domain-specific information.

Through the secondary extraction of the channel dimension by preserving or suppressing each channel in \(\mathbf{F}^{di}\) and \(\mathbf{F}^{ds}\), it has expanded the gap between domain-invariant features and domain-specific features, promoting the generalization ability of the model.

\subsection{Shift-Sensitive Adaptive Monitor}
\label{subsec: SSAM}

In cross-scene HSIs, domain shifts might be caused by various factors such as differences in imaging time, location, seasons, and sensors. Furthermore, the scale of domain shift between extracted features in source domain and target domain fluctuates dynamically with model training. Consequently, the scale of these domain shifts across different scenes and different training stages may differ greatly. Fixed domain alignment for different scenes and different training stages might induce negative transfer. Therefore, it is necessary to dynamically measure the scale of domain shift and adjust the alignment strategy, making it suitable for various scenes.

In SSAM, the scale of the domain shift is defined by  measuring the distribution of  the channel variance between source and target domains during the training process. Then, depending on the scale of the domain shift, the extent of alignment is dynamically adjusted by updating the mask ratio of domain-invariant features and domain-specific features.

Specifically,  \( \mu_e \)  is used to represent the scale of the domain shift. A larger \(\mu_e\) indicates a greater disparity of feature values across different domains, suggesting a larger domain shift and the need for a more aggressive alignment strategy. Conversely, a smaller value of \(\mu_e\) implies a lesser disparity in feature values between different domains, indicative of a smaller domain shift, and thus calling for a more gentle alignment strategy. Therefore, \(\mu_e\) is designed to update the mask ratio \(r_e\) for the alignment strategy. In order to map \(\mu_e\) into the \([0, 1]\) range, a shifted Sigmoid function is designed to establish the mapping. 

\begin{equation}
\label{eq: r1}
\begin{aligned}
    r'_e = \frac{1}{1+e^{-k (\mu_e - s) }},
\end{aligned}
\end{equation}
where \(r'_e\) is the temporary mask ratio for \(r_e\) of the \(e^{th}\) epoch, \(k\) and \(s\) represent the slope and offset adjustment parameters of the Sigmoid function, which are employed to yield a smooth output by mapping \( \mu_e \) to an appropriate range.

The calculation of \( \mu_e \) begins by measuring the channel variance between the source and target domains, and then computing the average of these variances, as shown below:

\begin{equation}
\begin{gathered}
    \mu_e = \frac{1}{C} \sum_{i=1}^{C} \Biggl(\frac{1}{n_s+n_t-1} \Bigl(\sum_{k=1}^{n_s} (\boldsymbol{P}^{di}_{s_k, c} - \overline{\boldsymbol{P}^{di}_c})^2 + \\
    \sum_{k=1}^{n_t} (\boldsymbol{P}^{di}_{t_k, c} - \overline{\boldsymbol{P}^{di}_c})^2 \Bigr)\Biggr),
\end{gathered}
\end{equation}
where \(n_s\) and \(n_t\) are the numbers of source samples and target samples, and \(\overline{\boldsymbol{P}^{di}_c}\) is the mean of channel variance, which is defined as follows:

\begin{equation}
\begin{gathered}
    \overline{\boldsymbol{P}^{di}_c} = \frac{1}{n_s+n_t} (\sum_{k=1}^{n_s} \boldsymbol{P}^{di}_{s_k, c} + \sum_{k=1}^{n_t} \boldsymbol{P}^{di}_{t_k, c}),
\end{gathered}
\end{equation}

To preserve the stability of the training process, the EMA method is employed to update the mask ratio \( r_e \) of the \(e^{th}\) epoch as
\begin{equation}
\label{eq: r2}
\begin{gathered}
    r_e = (1-m)\cdot r_{e-1}+m\cdot r'_e 
\end{gathered}.
\end{equation}

\subsection{Reversible Feature Extractor}

\begin{figure}[ht!]
    \centering
    \includegraphics[width=1.0\linewidth]{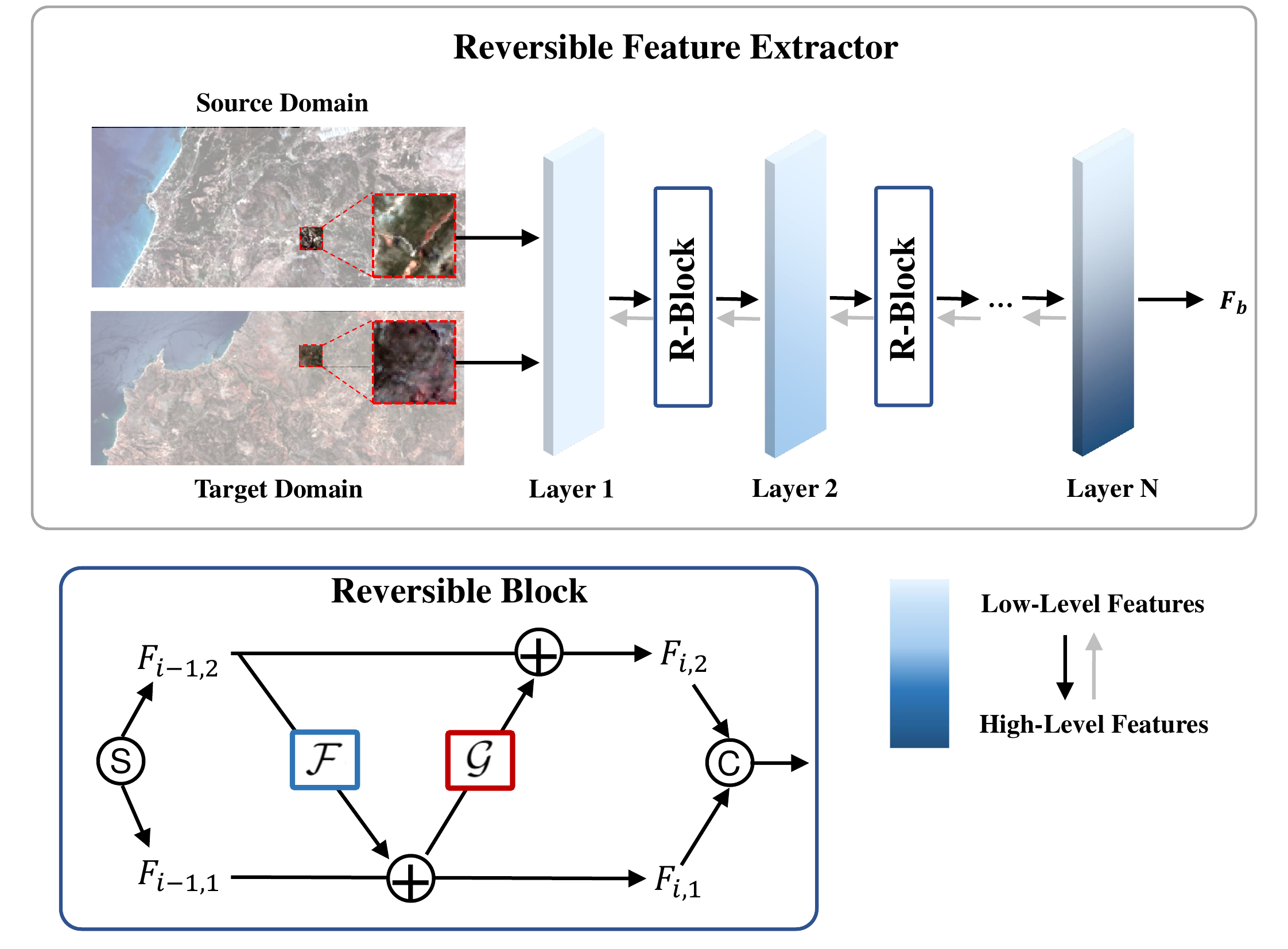}
    \caption{Model architecture of Reversible Feature Extractor. Light blue color represents low-level features and dark blue color represents high-level features.}
    \label{fig:reversible-feature-extractor}
\end{figure}

\begin{figure*}[ht!]
    \centering
    \subfloat[]{\includegraphics[width=0.48\linewidth]{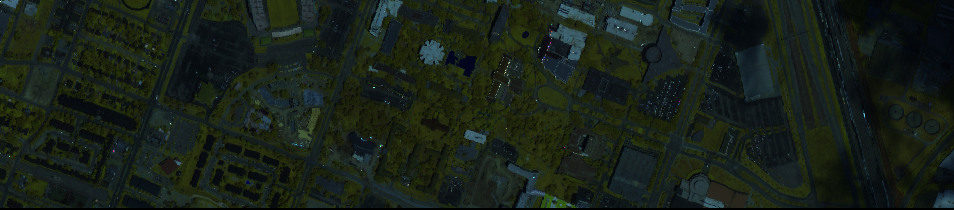}}
    \hfill
    \subfloat[]{\includegraphics[width=0.48\linewidth]{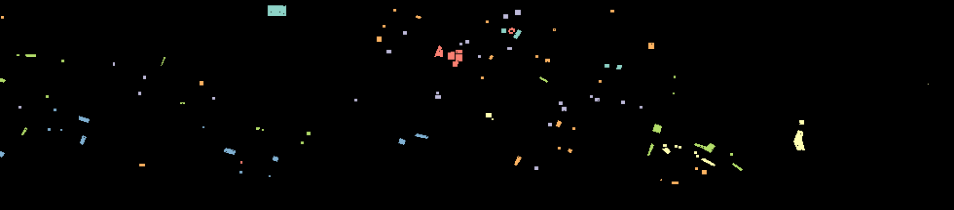}}
    \hfill
    \subfloat[]{\includegraphics[width=0.48\linewidth]{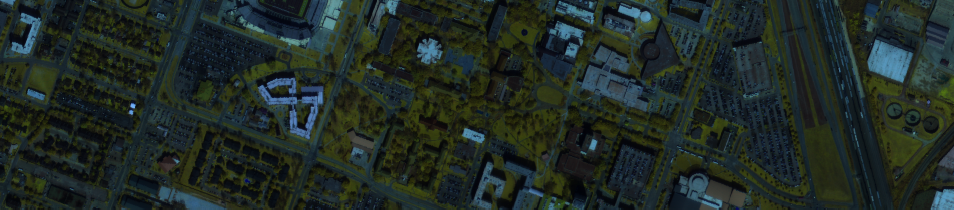}}
    \hfill
    \subfloat[]{\includegraphics[width=0.48\linewidth]{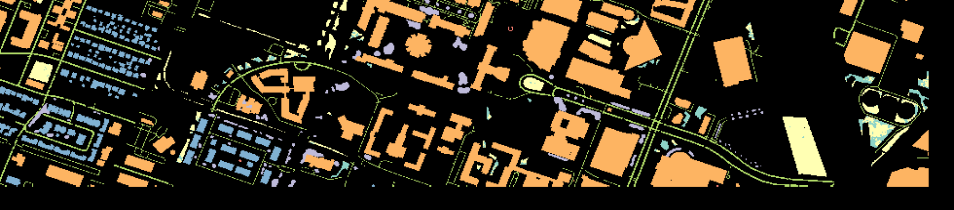}}
    \hfill
    \subfloat{\includegraphics[width=\linewidth]{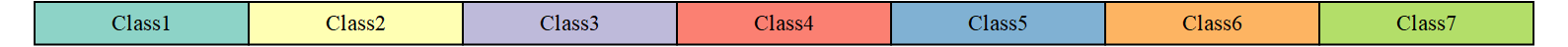}}
    \caption{The pseudocolor image and ground truth map of Houston dataset. (a) Pseudocolor image of Houston 2013. (b) Pseudocolor image of Houston 2018. (c) Ground-truth map of Houston 2013. (d) Ground-truth map of Houston 2018.}
    \label{fig:houston}
\end{figure*}
In HSIs, domain-invariant information and domain-specific information present in not just in high-level semantic features but also low-level features such as the corners of buildings and the texture and details of plants. However, existing methods primarily align high-level semantic features, neglecting the alignment of low-level features. For example, DAN~\cite{long2015learning} only aligns feature distribution in the last few layers, while DANN~\cite{ganin2015unsupervised, ganin2016domain} aligns the marginal feature distributions through adversarial training, and CDAN~\cite{long2018conditional} aligns joint distribution, which considers both feature representation and classifier prediction simultaneously. All these features are high-level features outputted by the feature extractor. These high-level features may lose low-level domain-invariant features as low-level features propagate layer by layer~\cite{zamir2018taskonomy, cai2022reversible} due to the Information Bottleneck principle~\cite{tishby2000information, tishby2015deep}.

To address this issue, the RFE is designed to retain low-level information, which is composed of numerous reversible layers. As shown in Fig.~\ref{fig:reversible-feature-extractor}, when RFE extracts high-level features represented by the blue color, the low-level features represented by the yellow color are not discarded but instead are embedded and disentangled into other dimensions. This is achieved through well-designed reversible layers. Owing to the reversibility of the layers, the process of forward propagation is lossless~\cite{gomez2017reversible}, ensuring the retention of low-level features while progressively extracting high-level features.

The input \(\boldsymbol{F}_{i-1}\) of reversible layers \(i\) (\(i \in \{2, 3, \ldots, n\}\)) is equally divided along the channel dimension into \((\boldsymbol{F}_{i-1, 1}, \boldsymbol{F}_{i-1, 2})\), and the corresponding outputs \(\boldsymbol{F}_{i}\) are \((\boldsymbol{F}_{i, 1}, F_{i, 2})\). The forward process is shown below.
\begin{equation}
\begin{aligned}
    \boldsymbol{F}_{i, 1} &= \boldsymbol{F}_{i-1, 1} + \mathcal{F}(\boldsymbol{F}_{i-1, 2}) \\
    \boldsymbol{F}_{i, 2} &= \boldsymbol{F}_{i-1, 2} + \mathcal{G}(\boldsymbol{F}_{i, 1}) \\
    \boldsymbol{F}^b &= \textrm{Concat}([\boldsymbol{F}_{n, 1}, \boldsymbol{F}_{n, 2}])
\end{aligned},
\end{equation}
where \(n\) is the number of reversible layers, and \(F^{b}\) is the output of the last reversible layer.

Correspondingly, the reverse process can reconstruct the inputs \((\boldsymbol{F}_{i-1, 1}, \boldsymbol{F}_{i-1, 2})\) from the outputs \((\boldsymbol{F}_{i, 1}, \boldsymbol{F}_{i, 2})\),
\begin{equation}
\begin{aligned}
    \boldsymbol{F}_{i-1, 2} &= \boldsymbol{F}_{i, 2} - \mathcal{G}(\boldsymbol{F}_{i, 1}) \\
    \boldsymbol{F}_{i-1, 1} &= \boldsymbol{F}_{i, 1} - \mathcal{F}(\boldsymbol{F}_{i-1, 2})
\end{aligned}.
\end{equation}
RFE not only accomplishes the extraction of high-level features but also embeds low-level features into other dimensions. Due to its rich feature information, it can provide more comprehensive domain-invariant features and domain-specific features during GSSD.

\subsection{Loss Function of S\texorpdfstring{\textsuperscript{4}}{4}DL}
S\textsuperscript{4}DL consists of three main components: RFE, GSSD and SSAM. All these components are updated by end-to-end training through Eq. \ref{eq:loss}. The overall loss \(\mathcal{L}_{\textrm{total}}\) of S\textsuperscript{4}DL is defined as follows: 

\begin{equation}
\label{eq:loss}
\begin{aligned}
    \mathcal{L}_{\textrm{total}} = \mathcal{L}_{\textrm{cls}} + \lambda_1\mathcal{L}_{\textrm{ortho}} + \lambda_2\mathcal{L}_{\textrm{dom}},
\end{aligned}
\end{equation}
where \(\mathcal{L}_{\textrm{cls}}\) is the cross-entropy loss for the labeled source domain~\cite{shannon2001mathematical}. \(\mathcal{L}_{\textrm{ortho}}\) is the orthogonal loss computed between \(\tilde{\mathbf{F}}^{di}\) and \(\tilde{\mathbf{F}}^{ds}\) to enhance their differentiation~\cite{bousmalis2016domain}. \(\mathcal{L}_{\textrm{dom}}\) is the domain classification loss~\cite{ganin2015unsupervised, ganin2016domain}, and \(\lambda_1\) and \(\lambda_2\) are hyperparameters that control the weight of the loss terms.

\section{Experimental Results and Analysis}
\label{sec:exp}

\subsection{Datasets}

\begin{table}
    \centering
    \caption{Classes and Numbers of Samples in Houston Dataset}
    \label{tab:houston}
    \begin{tabular}{c|c|c|c}
        \toprule[1.5pt]
        \multicolumn{2}{c|}{Class} & \multicolumn{2}{c}{Number of Samples} \\ 
        \midrule[1.0pt]
        No. & \multicolumn{1}{c|}{Name} & \makecell{Houston 2013\\(Source)} & \makecell{Houston 2018\\(Target)} \\ 
        \midrule[1.0pt]
        1 & Grass healthy & 345 & 1353 \\
        2 & Grass stressed & 365 & 4888 \\
        3 & Trees & 365 & 2766 \\
        4 & Water & 285 & 22 \\
        5 & Residential buildings & 319 & 5347 \\
        6 & Non-residential buildings & 408 & 32459 \\
        7 & Road & 443 & 6365 \\ 
        \midrule[1.0pt]
        \multicolumn{2}{c|}{Total} & 2530 & 53200 \\ 
        \bottomrule[1.5pt]
    \end{tabular}
\end{table}

\begin{table}
    \centering
    \caption{Classes and Numbers of Samples in HyRANK Dataset}
    \label{tab:hyrank}
    \begin{tabular}{c|c|c|c}
        \toprule[1.5pt]
        \multicolumn{2}{c|}{Class} & \multicolumn{2}{c}{Number of Samples} \\ 
        \midrule[1.0pt]
        No. & \multicolumn{1}{c|}{Name} & \makecell{Dioni\\(Source)} & \makecell{Loukia\\(Target)} \\ 
        \midrule[1.0pt]
        1 & Dense Urban Fabric & 1262 & 288 \\
        2 & Mineral Extraction Sites & 204 & 67 \\
        3 & Non Irrigated Arabel Land & 614 & 542 \\
        4 & Fruit Trees & 150 & 79 \\
        5 & Olive Groves & 1768 & 1401 \\
        6 & Coniferous Forest & 361 & 500 \\
        7 & Dense Sderophyllous Vegetation & 5035 & 3793 \\ 
        8 & Sparce Sderophyllous Vegetation & 6374 & 2803 \\
        9 & Sparcely Vegetated Area & 1754 & 404 \\
        10 & Rocks and Sand & 492 & 487 \\
        11 & Water & 1612 & 1393 \\
        12 & Coastal Water & 398 & 451 \\ 
        \midrule[1.0pt]
        \multicolumn{2}{c|}{Total} & 20024 & 12208 \\ 
        \bottomrule[1.5pt]
    \end{tabular}
\end{table}

\begin{table}
    \centering
    \caption{Classes and Numbers of Samples in S-H Dataset}
    \label{tab:s-h}
    \begin{tabular}{c|c|c|c}
        \toprule[1.5pt]
        \multicolumn{2}{c|}{Class} & \multicolumn{2}{c}{Number of Samples} \\ 
        \midrule[1.0pt]
        No. & \multicolumn{1}{c|}{Name} & \makecell{Hangzhou\\(Source)} & \makecell{Shanghai\\(Target)} \\ 
        \midrule[1.0pt]
        1 & Water & 18043 & 123123 \\
        2 & Land/Building & 77450 & 161689 \\
        3 & Plant & 40207 & 83188 \\
        \midrule[1.0pt]
        \multicolumn{2}{c|}{Total} & 135700 & 368000 \\ 
        \bottomrule[1.5pt]
    \end{tabular}
\end{table}

\begin{figure*}
    \centering
    \subfloat[]{\includegraphics[width=0.48\linewidth]{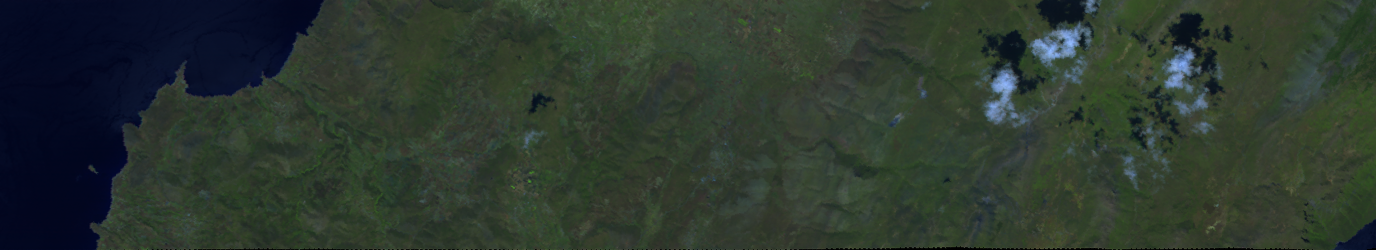}}
    \hfill
    \subfloat[]{\includegraphics[width=0.48\linewidth]{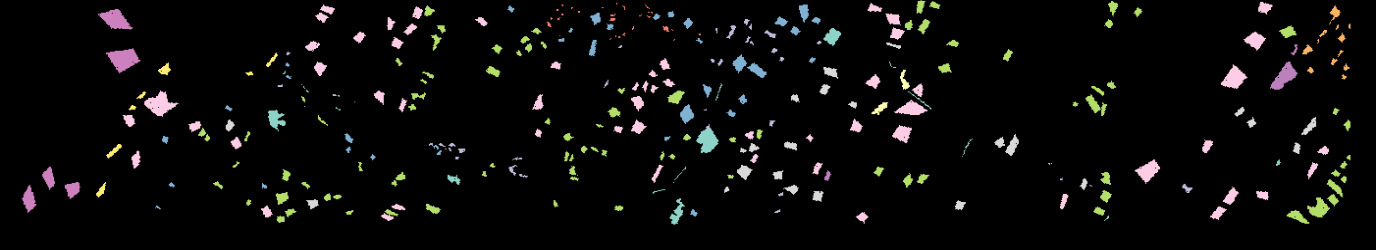}}
    \hfill
    \subfloat[]{\includegraphics[width=0.48\linewidth]{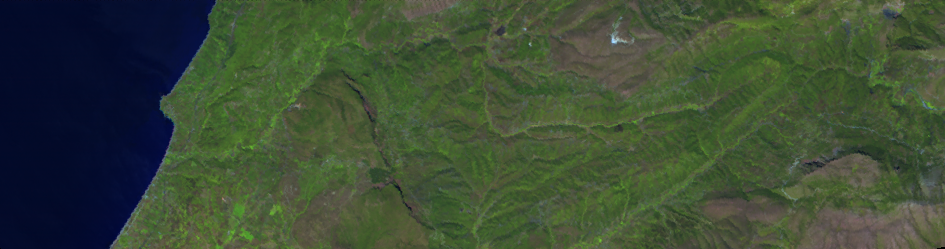}}
    \hfill
    \subfloat[]{\includegraphics[width=0.48\linewidth]{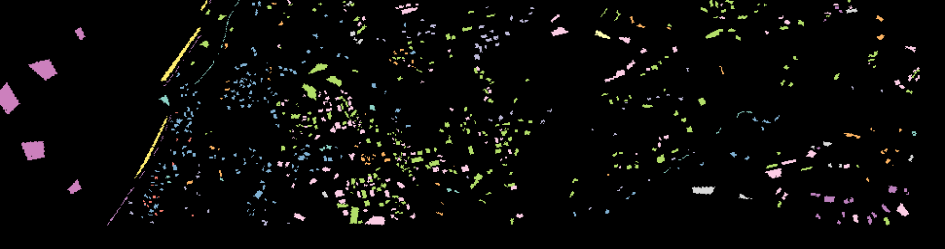}}
    \hfill
    \subfloat{\includegraphics[width=\linewidth]{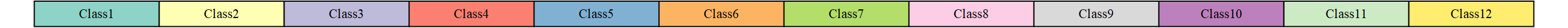}}
    \caption{The pseudocolor image and ground truth map of HyRANK dataset. (a) Pseudocolor image of Dioni. (b) Pseudocolor image of Loukia. (c) Ground-truth map of Dioni. (d) Ground-truth map of Loukia.}
    \label{fig:hyrank}
\end{figure*}

\begin{figure*}
    \centering
    \subfloat[]{\includegraphics[width=0.48\linewidth]{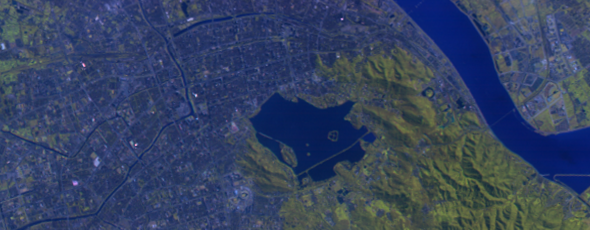}}
    \hfill
    \subfloat[]{\includegraphics[width=0.48\linewidth]{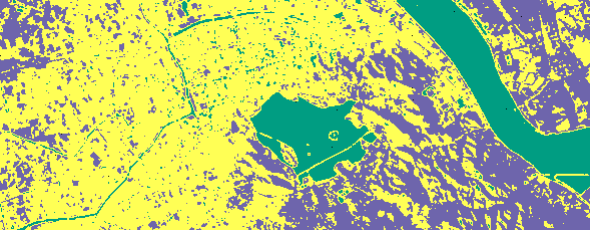}}
    \hfill
    \subfloat[]{\includegraphics[width=0.48\linewidth]{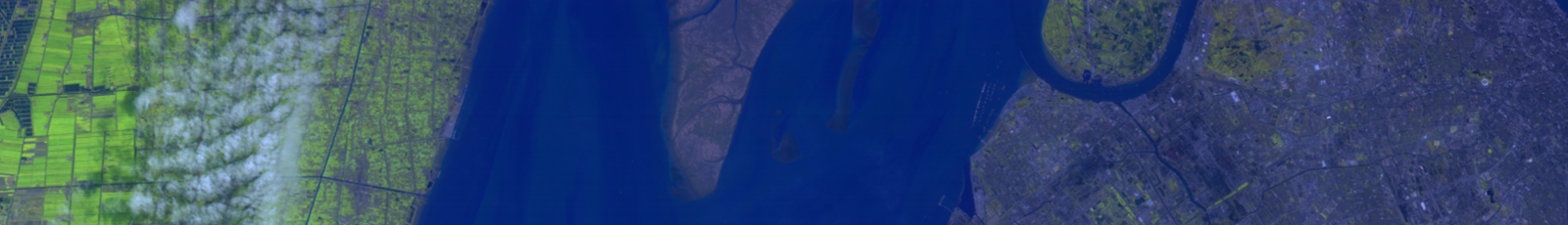}}
    \hfill
    \subfloat[]{\includegraphics[width=0.48\linewidth]{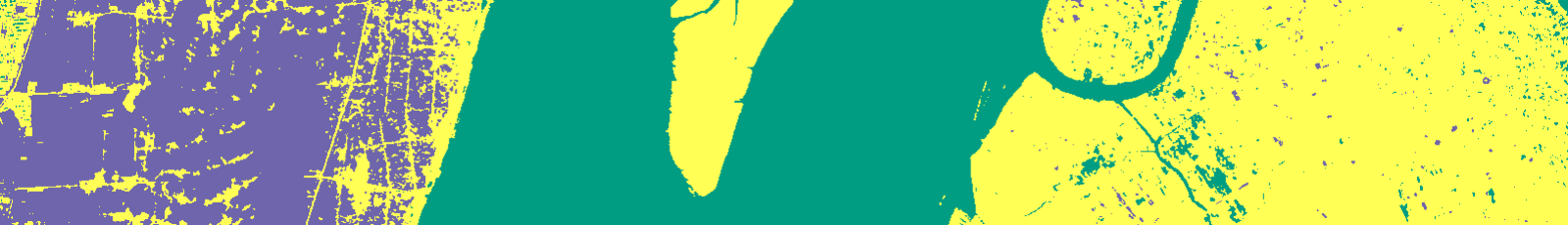}}
    \hfill
    \subfloat{\includegraphics[width=0.40\linewidth]{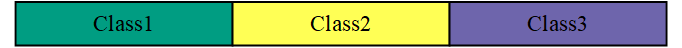}}
    \caption{The pseudocolor image and ground truth map of S-H dataset. (a) Pseudocolor image of Hangzhou. (b) Pseudocolor image of Shanghai. (c) Ground-truth map of Hangzhou. (d) Ground-truth map of Shanghai.}
    \label{fig:s-h}
\end{figure*}

\begin{table*}[ht!]
    \centering
    \caption{Class-specific and overall classification accuracy(\%) of different methods on Houston Dataset}
    \label{tab:res-houston}
    \footnotesize
    \begin{tabular}{c|ccccccc|cc}
    \toprule[1.5pt]
    Method & Class 1 & Class 2 & Class 3 & Class 4 & Class 5 & Class 6 & Class 7 & OA(\%) & Kappa(\%) \\
    
    \midrule[1.0pt]
    SVM        & 54.1$\pm$8.4  & 40.2$\pm$28.7 & 57.0$\pm$12.0 & 24.0$\pm$23.3  & 92.3$\pm$2.0  & 54.0$\pm$10.2 & 45.2$\pm$20.4 & 55.5$\pm$5.4  & 40.7$\pm$9.1  \\
    
    DDC~\cite{tzeng2014deep}      & 59.5$\pm$12.4 & 75.8$\pm$10.6 & 62.3$\pm$11.1 & 5.7$\pm$8.1   & 33.9$\pm$3.1  & 91.4$\pm$1.7  & 50.5$\pm$5.9  & 64.8$\pm$1.6  & 52.5$\pm$1.5  \\

    DAN~\cite{long2015learning}   & 63.4$\pm$8.3  & 84.6$\pm$5.9  & 62.4$\pm$6.9  & 4.4$\pm$2.9   & 30.5$\pm$1.4  & 90.8$\pm$2.5  & 49.5$\pm$6.3  & 63.4$\pm$1.3  & 51.0$\pm$1.3  \\

    JAN~\cite{long2017deep}       & 66.6$\pm$6.2  & 85.8$\pm$4.4  & 60.7$\pm$6.9  & 4.5$\pm$3.7   & 30.5$\pm$2.3  & 91.7$\pm$1.8  & 54.4$\pm$5.7  & 64.6$\pm$2.1  & 52.4$\pm$2.2  \\

    DSAN~\cite{zhu2020deep}       & 59.9$\pm$7.1  & 82.9$\pm$5.2  & 62.1$\pm$11.6 & 2.0$\pm$2.5   & 31.7$\pm$3.0  & 90.6$\pm$3.8  & 46.3$\pm$11.2 & 60.4$\pm$4.1  & 48.0$\pm$3.7  \\

    DANN~\cite{ganin2015unsupervised}       & 43.8$\pm$5.9  & 66.7$\pm$6.6  & 48.9$\pm$9.3  & 1.4$\pm$1.0    & 32.8$\pm$2.5  & 93.4$\pm$2.9  & 38.1$\pm$2.9  & 57.3$\pm$1.5  & 45.6$\pm$1.5  \\
    
    MCD~\cite{saito2018maximum}   & 40.3$\pm$7.1  & 52.2$\pm$7.6  & 49.0$\pm$6.1  & 1.1$\pm$0.7   & 35.4$\pm$5.4  & 90.9$\pm$3.1  & 45.6$\pm$14.5 & 58.2$\pm$3.3  & 45.7$\pm$2.9  \\

    ST~\cite{chen2022debiased}   & 47.9$\pm$9.6  & 69.6$\pm$6.1  & 48.5$\pm$6.2  & 0.8$\pm$0.3   & 30.4$\pm$3.0  & 94.8$\pm$1.5  & 40.2$\pm$7.7  & 56.0$\pm$3.3  & 44.6$\pm$3.0  \\

    DSN~\cite{bousmalis2016domain}& 51.8$\pm$12.6 & 52.8$\pm$7.3  & 62.0$\pm$11.7 & 3.1$\pm$5.4   & 32.1$\pm$3.3  & 93.2$\pm$3.2  & 47.0$\pm$9.1  & 60.1$\pm$2.8  & 48.0$\pm$2.6  \\
    
    SCLUDA~\cite{li2023supervised}& 58.4$\pm$14.7 & 77.5$\pm$5.5  & 50.8$\pm$6.8  & 50.9$\pm$18.1 & 85.8$\pm$4.2  & 62.9$\pm$2.9  & 47.4$\pm$9.4  & 64.0$\pm$2.7  & 48.4$\pm$3.4  \\
    
    TSTNet~\cite{zhang2021topological}  & 83.0$\pm$14.1 & 42.0$\pm$7.3  & 72.4$\pm$5.7  & 98.2$\pm$3.6   & 72.3$\pm$10.8 & 76.4$\pm$8.3  & 56.2$\pm$8.3  & 70.3$\pm$4.2  & 54.9$\pm$4.5  \\
   
    \midrule[1.0pt] 
    S\textsuperscript{4}DL (ours) & 74.9$\pm$6.7 & 72.1$\pm$7.2 & 73.8$\pm$13.5 & 18.9$\pm$18.7 & 36.4$\pm$3.9 & 93.1$\pm$1.2 & 67.5$\pm$9.1 & \textbf{72.1$\pm$2.3} & \textbf{60.7$\pm$2.6} \\    
    
    \bottomrule[1.5pt]
    \end{tabular}
\end{table*}

\begin{table*}[ht]
    \caption{Class-specific and overall classification accuracy(\%) of different methods on HyRANK Dataset}
    \label{tab:res-hyrank}
    \centering
    \footnotesize
    \resizebox{1.0\linewidth}{!}{
    \setlength{\tabcolsep}{1pt} 
    \centering
    \begin{tabular}{c|cccccccccccc|cc}
    \toprule[1.5pt]
    Method & Class 1 & Class 2 & Class 3 & Class 4 & Class 5 & Class 6 & Class 7 & Class 8 & Class 9 & Class 10 & Class 11 & Class 12 & OA(\%) & Kappa(\%) \\
    
    \midrule[1.0pt]
    SVM         & 19.3$\pm$8.4   & 100.0$\pm$0.0 & 0.2$\pm$0.6    & 0.0$\pm$0.0    & 47.4$\pm$1.4    & 0.2$\pm$0.6    & 47.4$\pm$1.4    & 45.0$\pm$2.8    & 15.0$\pm$5.9    & 0.1$\pm$0.3    & 100.0$\pm$0.0 & 84.6$\pm$7.3  & 41.6$\pm$0.5  & 38.6$\pm$0.6 \\
    
    DDC~\cite{tzeng2014deep} & 11.3$\pm$4.1   & 6.7$\pm$4.9             & 46.0$\pm$14.3  & 10.5$\pm$6.0   & 78.2$\pm$7.0    & 36.2$\pm$5.3   & 70.0$\pm$4.5    & 45.2$\pm$4.0    & 16.9$\pm$4.6    & 35.2$\pm$4.8   & 100.0$\pm$0.0 & 97.4$\pm$3.7  & 50.7$\pm$2.5  & 45.6$\pm$2.4 \\

    DAN~\cite{long2015learning} & 15.3$\pm$5.1   & 6.1$\pm$4.8             & 53.7$\pm$19.6  & 7.4$\pm$4.2    & 76.2$\pm$7.4    & 45.2$\pm$13.5  & 70.3$\pm$4.6    & 48.7$\pm$3.1    & 15.2$\pm$1.5    & 38.1$\pm$12.3  & 99.7$\pm$0.7  & 95.7$\pm$6.1  & 52.4$\pm$1.6  & 47.3$\pm$1.5 \\
    
    JAN~\cite{long2017deep} & 9.1$\pm$4.3    & 13.8$\pm$10.2           & 25.0$\pm$26.2  & 7.2$\pm$11.1   & 64.1$\pm$17.1   & 31.8$\pm$20.5  & 67.8$\pm$4.4    & 42.7$\pm$6.6    & 11.6$\pm$3.5    & 38.2$\pm$10.8  & 96.8$\pm$5.6  & 96.8$\pm$4.1  & 49.7$\pm$2.8  & 44.4$\pm$2.9 \\
    
    DSAN~\cite{zhu2020deep} & 4.6$\pm$3.3    & 5.1$\pm$12.7            & 1.9$\pm$5.6    & 0.0$\pm$0.0    & 29.6$\pm$18.0   & 7.5$\pm$17.8   & 67.0$\pm$4.8    & 41.4$\pm$6.6    & 7.9$\pm$4.9     & 18.5$\pm$17.9  & 82.1$\pm$8.9  & 0.0$\pm$0.0   & 47.4$\pm$2.4  & 41.4$\pm$2.3 \\

    DANN~\cite{ganin2015unsupervised} & 18.7$\pm$3.3    & 19.4$\pm$11.5           & 65.2$\pm$9.0   & 8.9$\pm$5.7    & 80.0$\pm$4.1    & 43.3$\pm$15.9  & 73.5$\pm$3.1    & 48.3$\pm$3.2    & 20.8$\pm$1.4    & 56.6$\pm$6.8   & 100.0$\pm$0.0 & 98.3$\pm$3.1  & 57.5$\pm$1.5  & 52.7$\pm$1.5 \\
    
    MCD~\cite{saito2018maximum} & 14.1$\pm$1.7   & 11.8$\pm$15.7           & 8.5$\pm$7.5    & 0.0$\pm$0.0    & 74.4$\pm$9.6    & 49.1$\pm$19.6  & 75.0$\pm$2.3    & 52.9$\pm$4.2    & 20.0$\pm$1.9    & 59.5$\pm$18.0  & 94.7$\pm$8.9  & 87.2$\pm$29.3 & 57.5$\pm$1.7  & 52.4$\pm$1.9 \\

    ST~\cite{zou2018unsupervised} & 7.1$\pm$7.1    & 14.1$\pm$30.1           & 0.0$\pm$0.0    & 0.0$\pm$0.0    & 40.2$\pm$21.6   & 0.0$\pm$0.0    & 61.4$\pm$6.5    & 52.1$\pm$8.2    & 21.1$\pm$12.5   & 44.2$\pm$32.5  & 74.9$\pm$9.7  & 0.0$\pm$0.0   & 53.2$\pm$3.8  & 47.0$\pm$3.8 \\
    
    DSN~\cite{bousmalis2016domain} & 24.1$\pm$5.7   & 8.7$\pm$9.8             & 67.8$\pm$10.2  & 27.9$\pm$8.9   & 86.6$\pm$1.6    & 57.9$\pm$13.6 & 75.4$\pm$2.6    & 50.1$\pm$1.6    & 24.1$\pm$2.0    & 68.5$\pm$10.3  & 98.8$\pm$2.8  & 89.4$\pm$29.8 & 61.0$\pm$1.4  & 56.2$\pm$1.5 \\
    
    SCLUDA~\cite{li2023supervised} & 46.9$\pm$6.9 & 96.7$\pm$6.6           & 9.5$\pm$8.4    & 0.0$\pm$0.0    & 19.3$\pm$9.4    & 1.5$\pm$0.6    & 51.6$\pm$3.5    & 51.6$\pm$3.5    & 79.9$\pm$15.1 & 33.5$\pm$29.4 & 100.0$\pm$0.0 & 33.5$\pm$29.4 & 52.2$\pm$1.8  & 49.5$\pm$2.1 \\
    
    TSTNet~\cite{zhang2021topological} & 30.6$\pm$17.7 & 0.0$\pm$0.0             & 27.3$\pm$25.0  & 0.0$\pm$0.0    & 69.2$\pm$12.3   & 4.5$\pm$6.2    & 75.7$\pm$6.8    & 54.1$\pm$3.7 & 80.4$\pm$6.7 & 2.1$\pm$3.7   & 100.0$\pm$0.0 & 80.0$\pm$40.0 & 63.1$\pm$1.6  & 54.9$\pm$1.9 \\
     
     \midrule[1.0pt]
    S\textsuperscript{4}DL (ours) & 29.3$\pm$4.8 & 9.3$\pm$20.4 & 70.2$\pm$11.1 & 33.9$\pm$10.0 & 91.9$\pm$1.5 & 44.2$\pm$27.5 & 76.8$\pm$2.6 & 53.1$\pm$2.4 & 30.5$\pm$3.1 & 70.2$\pm$13.4 & 99.6$\pm$0.6 & 89.9$\pm$30.0 & \textbf{65.0$\pm$1.9} & \textbf{60.2$\pm$2.0} \\
    
    \bottomrule[1.5pt]
    \end{tabular}}
\end{table*}

\begin{table*}[ht]
    \caption{Class-specific and overall classification accuracy(\%) of different methods on S-H Dataset}
    \label{tab:res-sh}
    \footnotesize 
    \centering
    \begin{tabular}{c|ccc|cc}
    \toprule[1.5pt]
    Method & Class 1 & Class 2 & Class 3 & OA(\%) & Kappa(\%) \\
    
    \midrule[1.0pt]
    SVM         & 100.0$\pm$0.0 & 74.0$\pm$15.0 & 82.7$\pm$1.7 & 81.1$\pm$9.4 & 74.2$\pm$11.2 \\
    
    DDC~\cite{tzeng2014deep} & 100.0$\pm$0.0 & 85.9$\pm$9.0 & 83.1$\pm$2.7 & 88.4$\pm$5.4 & 83.3$\pm$7.2 \\

    DAN~\cite{long2015learning} & 100.0$\pm$0.0 & 79.9$\pm$15.5 & 86.7$\pm$5.0 & 83.9$\pm$10.8 & 77.7$\pm$14.0 \\
    
    JAN~\cite{long2017deep} & 100.0$\pm$0.0 & 86.7$\pm$7.7 & 85.2$\pm$4.1 & 89.5$\pm$4.9 & 84.8$\pm$6.7 \\
        
    DSAN~\cite{zhu2020deep} & 100.0$\pm$0.0 & 84.7$\pm$11.0 & 81.5$\pm$10.5 & 87.2$\pm$9.2 & 82.1$\pm$11.6 \\
    
    DANN~\cite{ganin2015unsupervised} & 100.0$\pm$0.0 & 77.7$\pm$6.7 & 85.9$\pm$9.0 & 84.2$\pm$3.8 & 77.5$\pm$5.1 \\
       
    MCD~\cite{saito2018maximum} & 100.0$\pm$0.0 & 89.5$\pm$2.9 & 86.7$\pm$4.6 & 91.7$\pm$1.9 & 87.7$\pm$2.6 \\

    ST~\cite{zou2018unsupervised} & 100.0$\pm$0.0 & 85.6$\pm$3.7 & 88.7$\pm$5.1 & 90.0$\pm$1.8 & 85.4$\pm$2.6 \\
    
    DSN~\cite{bousmalis2016domain} & 100.0$\pm$0.0 & 91.5$\pm$1.9 & 80.8$\pm$4.0 & 91.1$\pm$2.0 & 87.0$\pm$2.8 \\
    
    SCLUDA~\cite{li2023supervised} & 89.8$\pm$1.5 & 89.0$\pm$1.1 & 98.5$\pm$0.3 & 91.4$\pm$0.8 & 86.8$\pm$1.3 \\
    
    TSTNet~\cite{zhang2021topological} & 86.2$\pm$4.8 & 68.8$\pm$3.1 & 100.0$\pm$0.0 & 81.7$\pm$2.2 & 72.6$\pm$3.2 \\
    
    \midrule[1.0pt]
    S\textsuperscript{4}DL (ours) & 100.0$\pm$0.0 & 91.1$\pm$2.7 & 86.4$\pm$2.3 & \textbf{92.4$\pm$1.2} & \textbf{88.8$\pm$1.7} \\

    \bottomrule[1.5pt]
    \end{tabular}
\end{table*}

For performance evaluation, three challenging HSI datasets, Houston, HyRANK and S-H, are selected, and the performance on these datasets is examined in terms of class-specific accuracy, overall accuracy (OA), and Kappa coefficient.

\textbf{Houston.}
The Houston dataset is compose of Houston-2013~\cite{debes2014hyperspectral} and Houston-2018~\cite{le20182018}, captured by different sensors in 2013 and 2018 over the University of Houston, Texas, USA. 
Houston-2013 contains $349\times1905$ pixels with 144 spectral bands at a spatial resolution of 2.5 meters, and the Houston-2018 dataset contains $210\times954$ pixels with 48 spectral bands, offering a finer spatial resolution of 1 meter. 
The overlapped 48 spectral bands are collected from both images. 
Following~\cite{zhang2021topological}, 210×954 pixels from Houston 2013 are selected as the source domain, and Houston 2018 is used as the target domain. 
Pixel-wise annotations of 7 categories are provided for both images, as detailed in Table .~\ref{tab:houston} and visualized in Fig.~\ref{fig:houston}.

\textbf{HyRANK.}
The HyRANK dataset~\cite{karantzalos2018hyrank} covers two hyperspectral scenes, Dioni and Loukia. 
Both of them are captured by the EO-1 Hyperion hyperspectral sensor.  
The source domain, Dioni, consists of 250×1376 pixels and 176 bands, and the target domain, Loukia, comprises 249×945 pixels and 176 bands. 
The annotations for 12 categories are provided, and please see Table \ref{tab:hyrank} for more details on the number of samples of these categories. Fig.~\ref{fig:hyrank} presents the pseudo-color images and their corresponding ground truth maps.

\textbf{S-H.}
The Shanghai-Hangzhou dataset was acquired using the EO-1 Hyperion hyperspectral sensor, which features 220 spectral bands.
The source domain, the Hangzhou scene, comprises 590×360 pixels, while the target domain, the Shanghai scene, includes 1660×260 pixels. 
After the removal of bad bands \cite{zhang2021topological}, 198 bands are remained. 
The annotations on three categories of land covers are provided, which are Water, Land/Building and Plant. 
Table \ref{tab:s-h} summarizes the number of samples and a visualization on the images and their corresponding ground truth maps are provided in Fig.~\ref{fig:s-h}.

\subsection{Implementation Details}
\label{sec:exp-setting}
 
For a fair comparison, the input patch size is set as 11 × 11 for all the methods, and Z-score normalization is conducted prior to putting the data into the network. Adaptive moment estimation(Adam) is utilized as the optimization scheme. We adopted a plateau strategy for learning rate decay, applying a decay factor of 0.1 and a patience of 2. 
In S\textsuperscript{4}DL, the offset \(k\) of the shifted Sigmoid function is 1.5, and the slope \(s\) is 2.5 on three datasets, which will be discussed in detail in Section \ref{sec: pt}.
All the models were trained 10 times using different random seeds, and the averaged results are recorded. All the experiments were conducted by PyTorch 2.0 on NVIDIA GeForce RTX 3090 GPU.

\subsection{Main Results}

For validating the effectiveness of our S\textsuperscript{4}DL, a Support Vector Machine (SVM) baseline without any domain adaptation and 10 top-performing UDA methods are selected for comparison. 
On all datasets, we collect the average and variance of the reported evaluation metrics from 10 rounds of experiments by each method.

Among the selected UDA methods, DDC~\cite{tzeng2014deep}, DAN~\cite{long2015learning}, JAN~\cite{long2017deep} and DSAN~\cite{zhu2020deep} are statistics matching methods, where both DDC and DAN use Maximum Mean Discrepancy~\cite{gretton2012kernel} loss for adaptation, JAN uses Joint Maximum Mean Discrepancy loss, and DSAN uses Local Maximum Mean Discrepancy loss, with the number of kernels of DAN, JAN, and DSAN being 5. 
DANN~\cite{ganin2015unsupervised} and MCD~\cite{saito2018maximum} are domain adversarial methods that share a discriminator architecture identical to S\textsuperscript{4}DL. 
ST~\cite{zou2018unsupervised}, a semi-supervised method, operates with a confidence threshold set to 0.7. 
DSN~\cite{bousmalis2016domain} is a domain disentangling method, utilizing a uniform backbone for the shared encoder, the private target encoder, and the private source encoder. 
SCLUDA~\cite{li2023supervised} and TSTNet~\cite{zhang2021topological} are recent corss-scene HSI classification methods, and we reproduce the results by following their original setups.

\newcommand{\resultvspace}{3pt}

\begin{figure*}[ht]
\centering
\resizebox{\linewidth}{!}{%
\subfloat[]{\rotatebox{-90}{\includegraphics[width=0.5\linewidth]{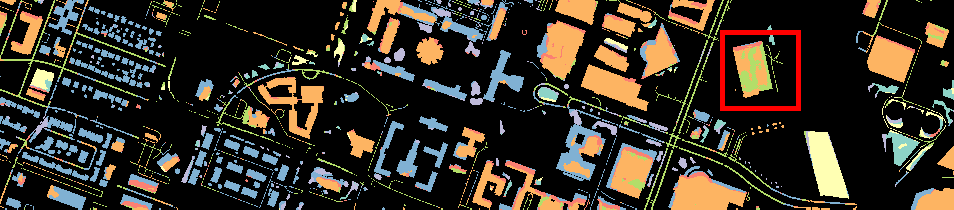}}%
\label{result_houston_nommd}}
\vspace{\resultvspace}
\subfloat[]{\rotatebox{-90}{\includegraphics[width=0.5\linewidth]{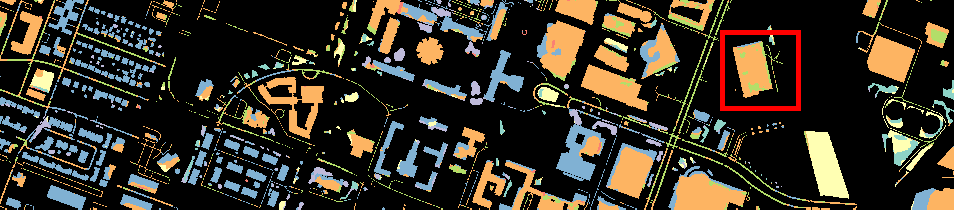}}%
\label{result_houston_ddc}}
\vspace{\resultvspace}
\subfloat[]{\rotatebox{-90}{\includegraphics[width=0.5\linewidth]{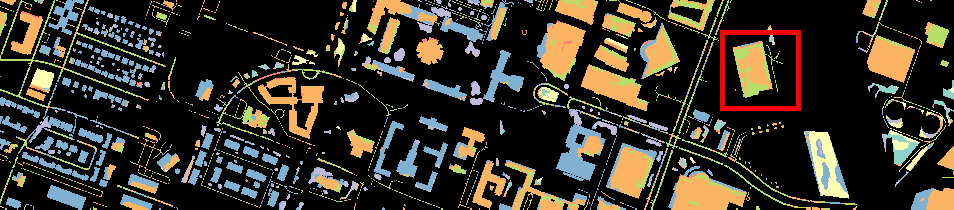}}%
\label{result_houston_dan}}
\vspace{\resultvspace}
\subfloat[]{\rotatebox{-90}{\includegraphics[width=0.5\linewidth]{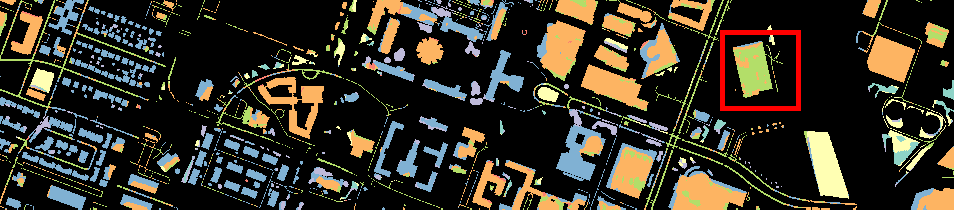}}%
\label{result_houston_jan}}
\vspace{\resultvspace}
\subfloat[]{\rotatebox{-90}{\includegraphics[width=0.5\linewidth]{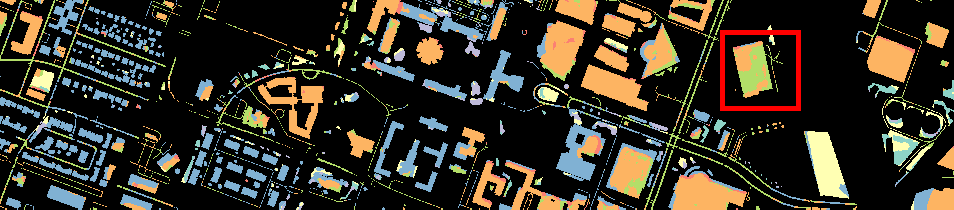}}%
\label{result_houston_dsan}}
\vspace{\resultvspace}
\subfloat[]{\rotatebox{-90}{\includegraphics[width=0.5\linewidth]{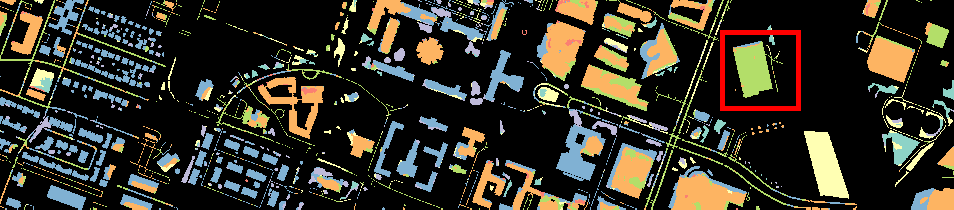}}%
\label{result_houston_dann}}
\vspace{\resultvspace}
\subfloat[]{\rotatebox{-90}{\includegraphics[width=0.5\linewidth]{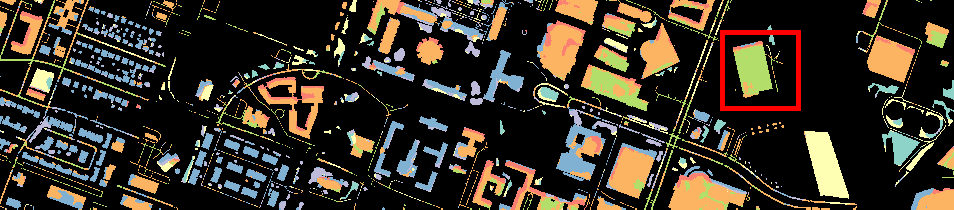}}%
\label{result_houston_mcd}}
\vspace{\resultvspace}
\subfloat[]{\rotatebox{-90}{\includegraphics[width=0.5\linewidth]{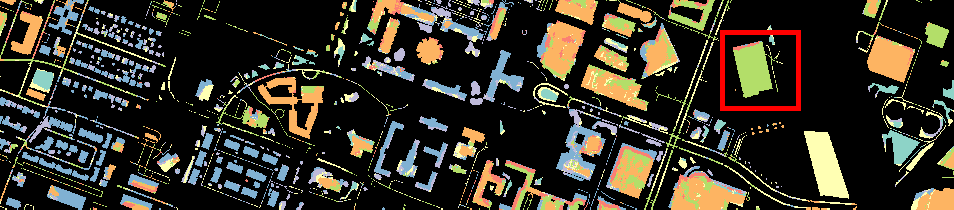}}%
\label{result_houston_dst}}
\vspace{\resultvspace}
\subfloat[]{\rotatebox{-90}{\includegraphics[width=0.5\linewidth]{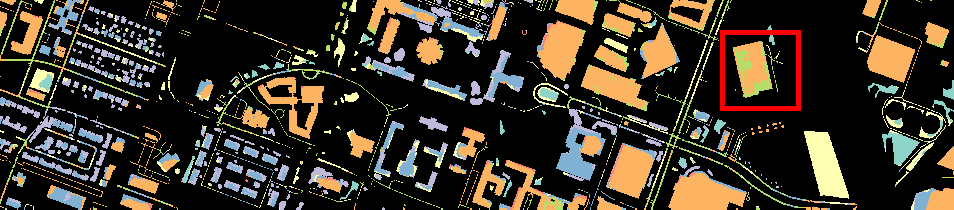}}%
\label{result_houston_dsn}}
\vspace{\resultvspace}
\subfloat[]{\rotatebox{-90}{\includegraphics[width=0.5\linewidth]{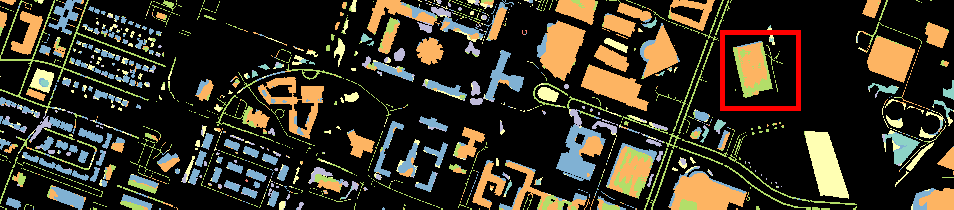}}%
\label{result_houston_scluda}}
\vspace{\resultvspace}
\subfloat[]{\rotatebox{-90}{\includegraphics[width=0.5\linewidth]{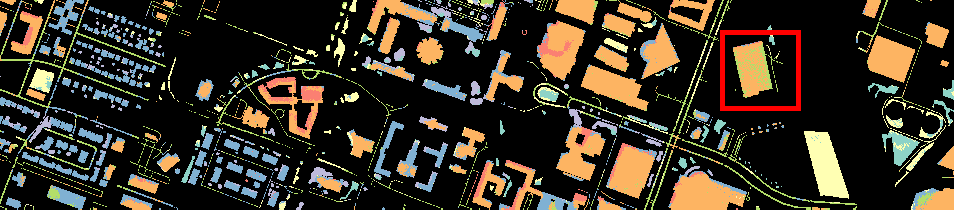}}%
\label{result_houston_tstnet}
}
\vspace{\resultvspace}
\subfloat[]{\rotatebox{-90}{\includegraphics[width=0.5\linewidth]{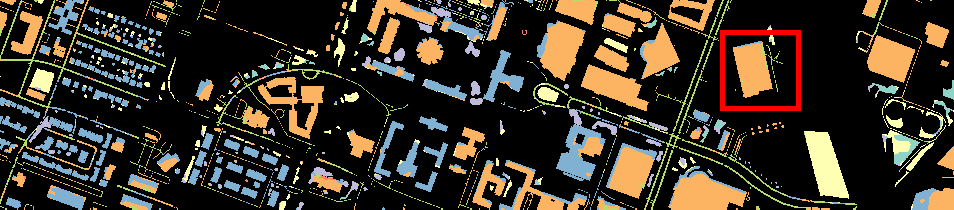}}%
\label{fig:result_houston_our}
}
}
\caption{Classification map for Houston with different methods: (a) SVM; (b) DDC; (c) DAN; (d) JAN; (e) DSAN; (f) DANN; (g) MCD;(h) ST; (i) DSN; (j) SCLUDA; (k) TSTNet and (l) S\textsuperscript{4}DL.}
\label{fig:res_vis_houston}
\end{figure*}

\textbf{Houston.}
As summarized in Table.~\ref{tab:houston}, the proposed S\textsuperscript{4}DL achieves the highest OA and Kappa scores. 
More specifically, our S\textsuperscript{4}DL outperforms the existing top-performer TSTNet by 1.8\% and 5.8\% in term of OA and Kappa, respectively.
Notably, on the categories of Grass stressed, Trees, Non-residential buildings and Road, the accuracy has increased by 30.1\%, 1.4\%, 16.7\% and 11.3\%, respectively.
Compared with DSN that is based representation disentangling without channel decomposition, the proposed S\textsuperscript{4}DL is higher by 12.0\% in OA and 12.7\% in Kappa scores owing to the adaptive disentangling strategy in channel dimensions.
Qualitatively, with enhanced ability to capture domain-invariant features, our S\textsuperscript{4}DL tends to generalize better to unseen scenes and produce classification with reduced false alarms.
As visualized in Fig.~\ref{fig:res_vis_houston}, our S\textsuperscript{4}DL exhibits more precise classification on Non-residential buildings located at the bottom of the image, with significantly reduced noise. 
The superior performance on this dataset implies that S\textsuperscript{4}DL demonstrates a robust capability in capturing domain-invariant information during the training process.

\begin{figure*}[t]
\centering
\resizebox{\linewidth}{!}{%
\subfloat[]{\rotatebox{-90}{\includegraphics[width=0.5\linewidth]{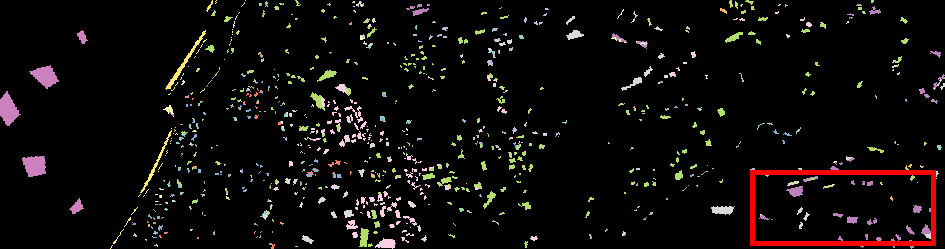}}%
\label{result_hyrank_nommd}}
\vspace{\resultvspace}
\subfloat[]{\rotatebox{-90}{\includegraphics[width=0.5\linewidth]{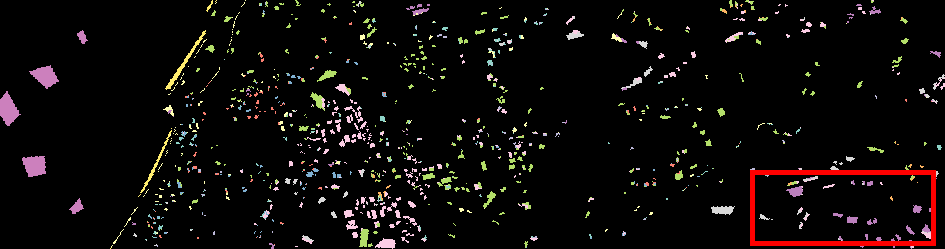}}%
\label{result_hyrank_ddc}}
\vspace{\resultvspace}
\subfloat[]{\rotatebox{-90}{\includegraphics[width=0.5\linewidth]{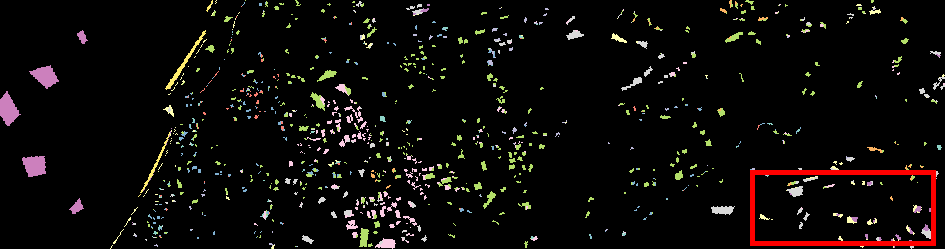}}%
\label{result_hyrank_dan}}
\vspace{\resultvspace}
\subfloat[]{\rotatebox{-90}{\includegraphics[width=0.5\linewidth]{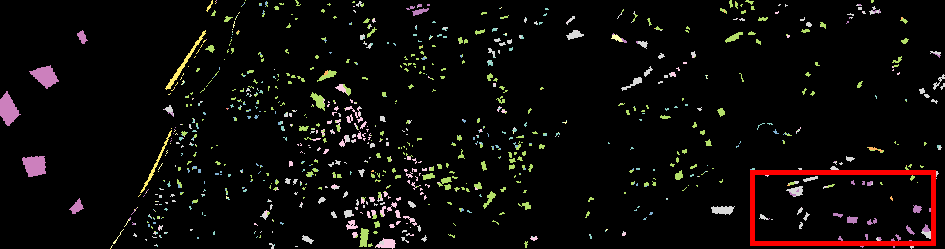}}%
\label{result_hyrank_jan}}
\vspace{\resultvspace}
\subfloat[]{\rotatebox{-90}{\includegraphics[width=0.5\linewidth]{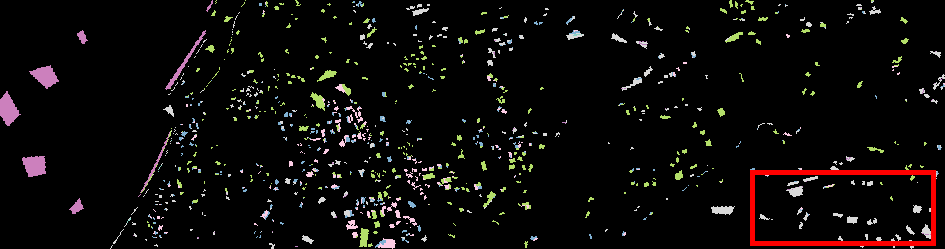}}%
\label{result_hyrank_dsan}}
\vspace{\resultvspace}
\subfloat[]{\rotatebox{-90}{\includegraphics[width=0.5\linewidth]{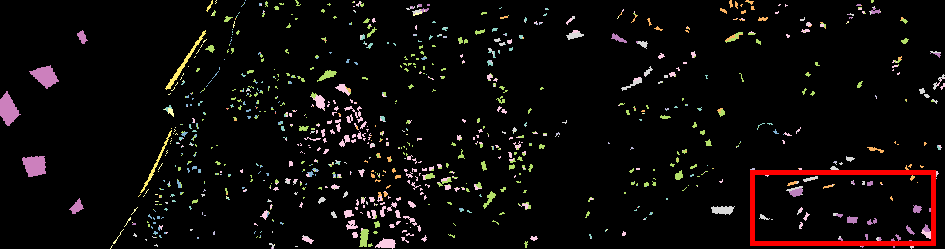}}%
\label{result_hyrank_dann}}
\vspace{\resultvspace}
\subfloat[]{\rotatebox{-90}{\includegraphics[width=0.5\linewidth]{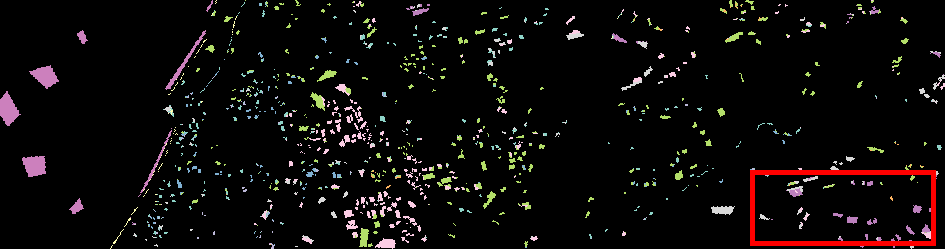}}%
\label{result_hyrank_mcd}}
\vspace{\resultvspace}
\subfloat[]{\rotatebox{-90}{\includegraphics[width=0.5\linewidth]{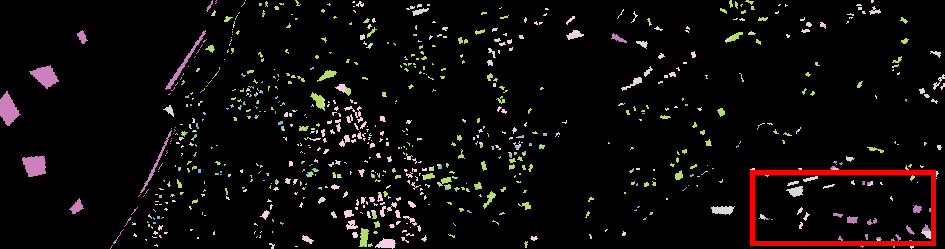}}%
\label{result_hyrank_dst}}
\vspace{\resultvspace}
\subfloat[]{\rotatebox{-90}{\includegraphics[width=0.5\linewidth]{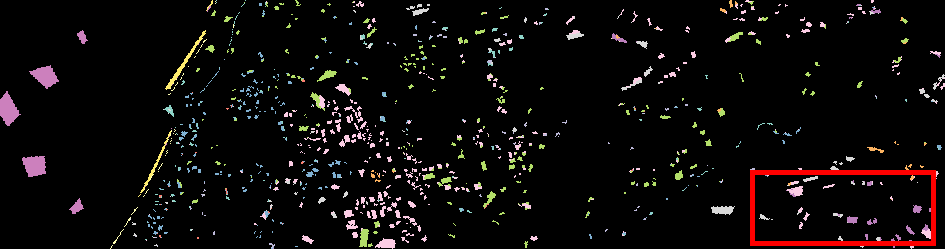}}%
\label{result_hyrank_dsn}}
\vspace{\resultvspace}
\subfloat[]{\rotatebox{-90}{\includegraphics[width=0.5\linewidth]{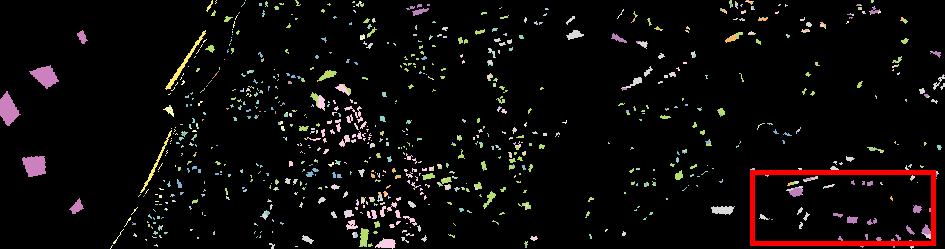}}%
\label{result_hyrank_scluda}}
\vspace{\resultvspace}
\subfloat[]{\rotatebox{-90}{\includegraphics[width=0.5\linewidth]{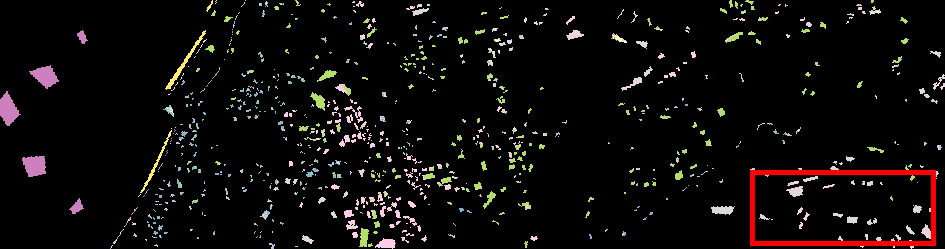}}%
\label{result_hyrank_tstnet}
}
\vspace{\resultvspace}
\subfloat[]{\rotatebox{-90}{\includegraphics[width=0.5\linewidth]{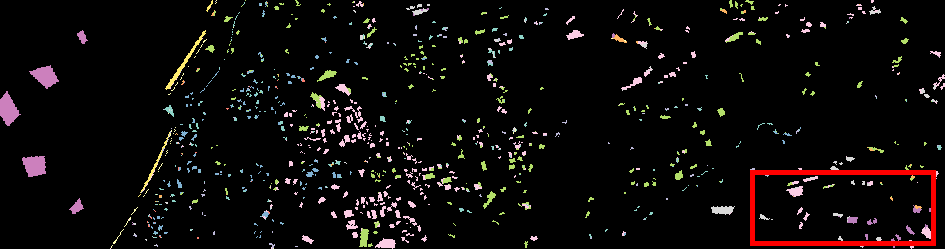}}%
\label{result_hyrank_our}
}
}
\caption{Classification map for HyRANK with different methods: (a) SVM; (b) DDC; (c) DAN; (d) JAN; (e) DSAN; (f) DANN; (g) MCD;(h) ST; (i) DSN; (j) SCLUDA; (k) TSTNet and (l) S\textsuperscript{4}DL.}
\label{fig:res_vis_hyrank}
\end{figure*}

\textbf{HyRANK.}
Table.~\ref{tab:hyrank} shows the cross-scene classification results in HyRANK dataset. 
Compared with suboptimal TSTNet, the proposed S\textsuperscript{4}DL has improved by 1.9\% in OA and 5.3\% in Kappa scores. 
Compared with the third best DSN, our S\textsuperscript{4}DL has improved by 4.0\% in OA and 4.0\% in Kappa scores, which confirm that our method can separate domain-invariant and domain-specific features more comprehensively, thereby further aiding the learning of discriminative features. 
Meanwhile, among all DA methods, our S\textsuperscript{4}DL achieves the best results on most 12 categories. In addition, it is difficult for most algorithms to correctly classify the Fruit trees and Rocks and Sand, while our S\textsuperscript{4}DL improves these categories by up to 33.9\% and 70.1\%, and by at least 6.0\% and 1.7\% respectively. 
From Fig.~\ref{fig:res_vis_hyrank}, it can be observed that our S\textsuperscript{4}DL successfully differentiates between easily confused Sparse Sclerophyllous Vegetation and Rocks and Sand in the lower left corner. This correct classification by our S\textsuperscript{4}DL in an area with significant inter-domain and minimal inter-class differences demonstrates its ability to effectively extract key discriminative features through suitable domain alignment.

\begin{figure*}
\centering
\resizebox{\linewidth}{!}{%
\subfloat[]{\includegraphics[width=0.1\linewidth]{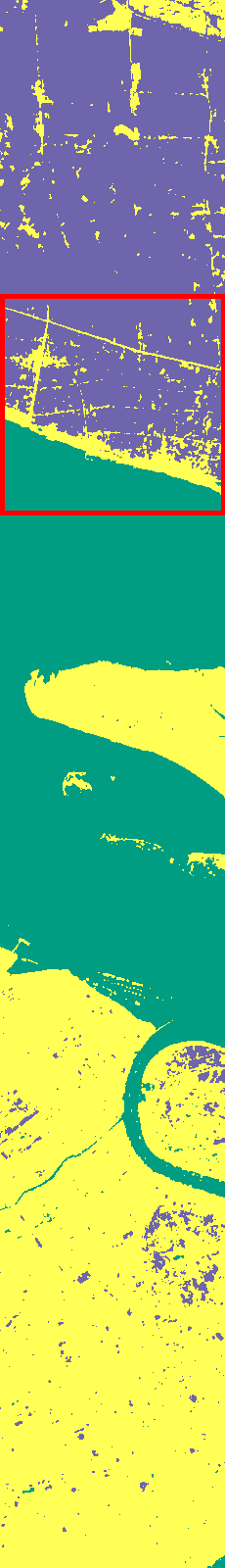}}%
\label{results_hanghang_nommd}
\vspace{\resultvspace}
\subfloat[]{\includegraphics[width=0.1\linewidth]{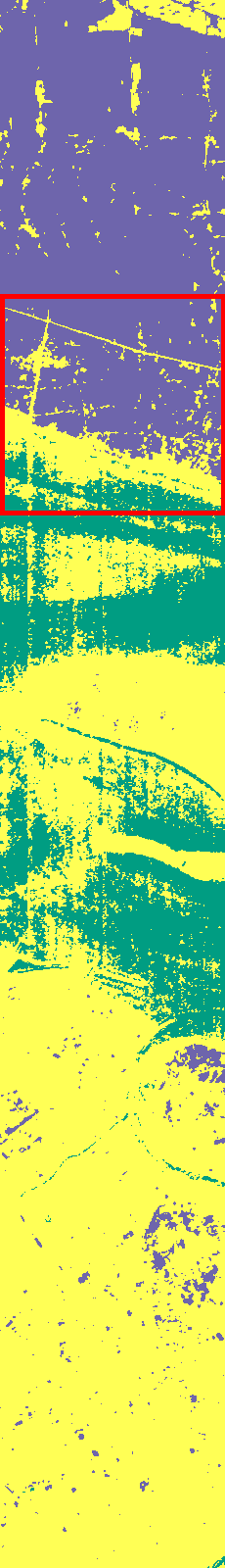}}%
\label{result_shanghang_ddc}
\vspace{\resultvspace}
\subfloat[]{\includegraphics[width=0.1\linewidth]{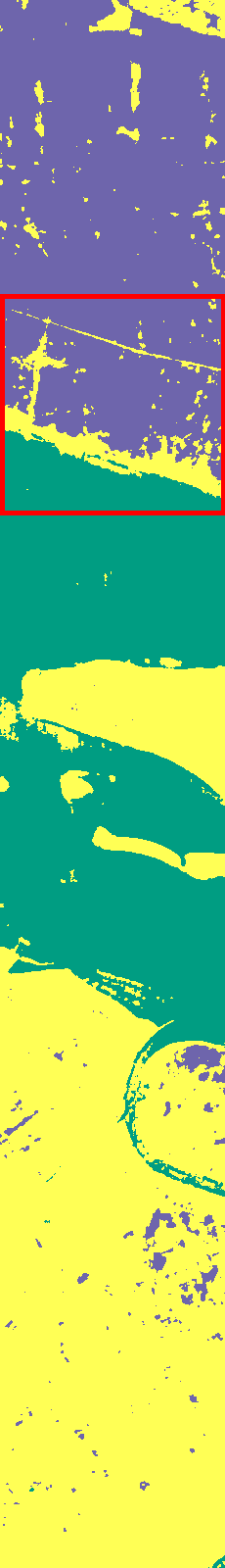}}%
\label{result_shanghang_dan}
\vspace{\resultvspace}
\subfloat[]{\includegraphics[width=0.1\linewidth]{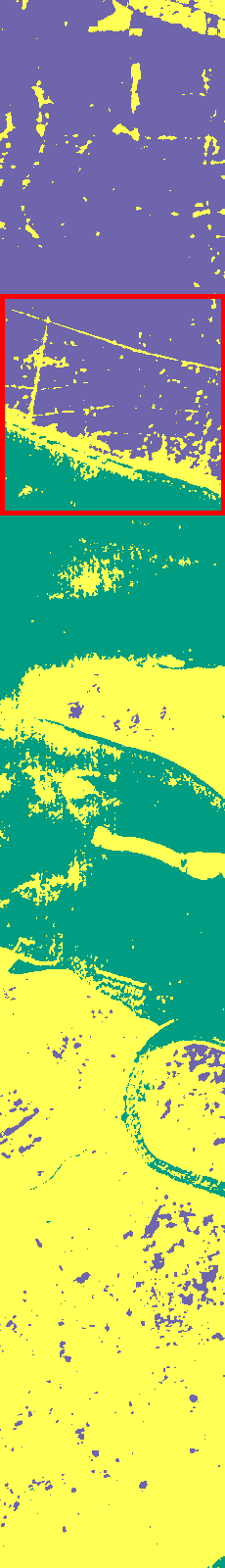}}%
\label{result_shanghang_jan}
\vspace{\resultvspace}
\subfloat[]{\includegraphics[width=0.1\linewidth]{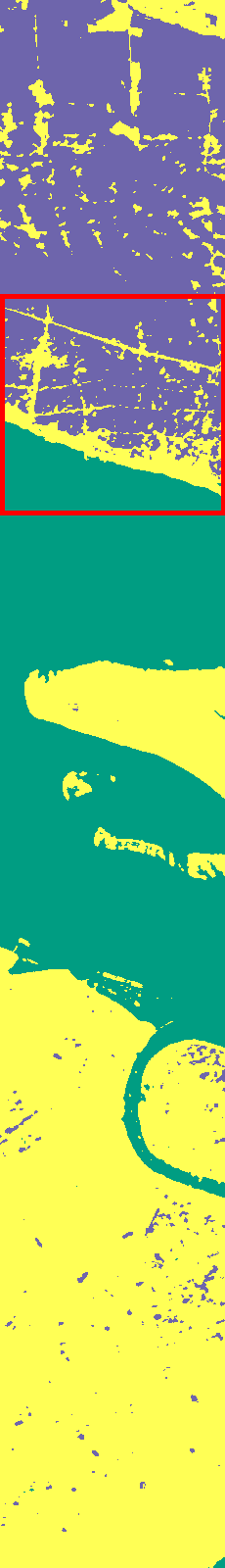}}%
\label{result_shanghang_dsan}
\vspace{\resultvspace}
\subfloat[]{\includegraphics[width=0.1\linewidth]{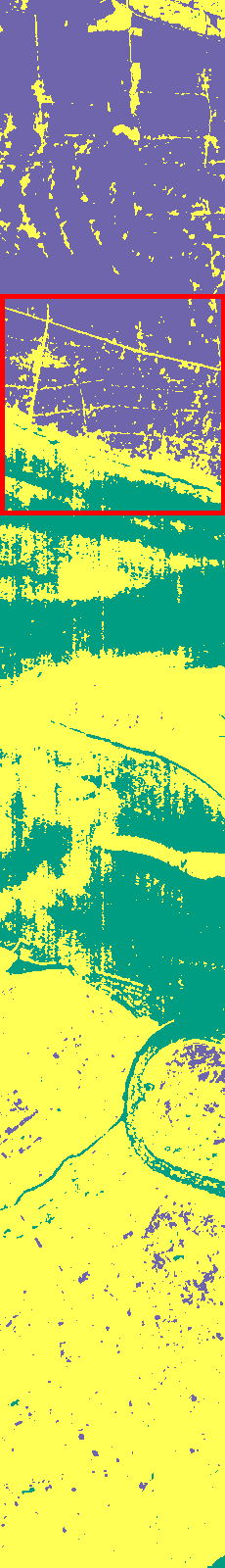}}%
\label{result_shanghang_dann}
\vspace{\resultvspace}
\subfloat[]{\includegraphics[width=0.1\linewidth]{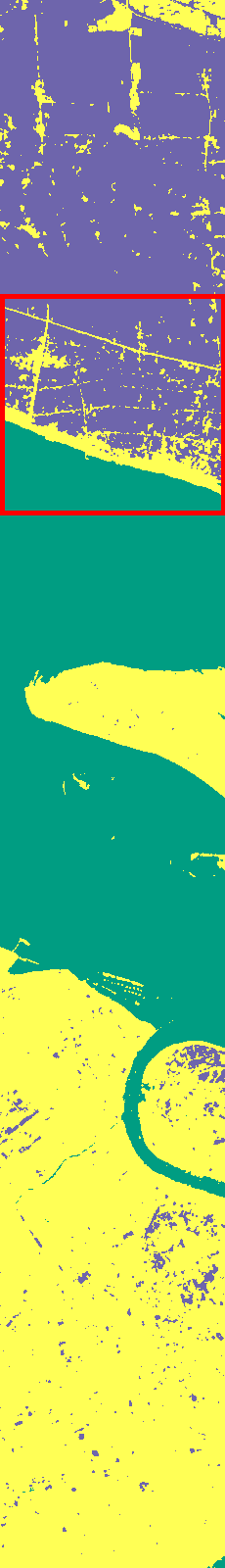}}%
\label{result_shanghang_mcd}
\vspace{\resultvspace}
\subfloat[]{\includegraphics[width=0.1\linewidth]{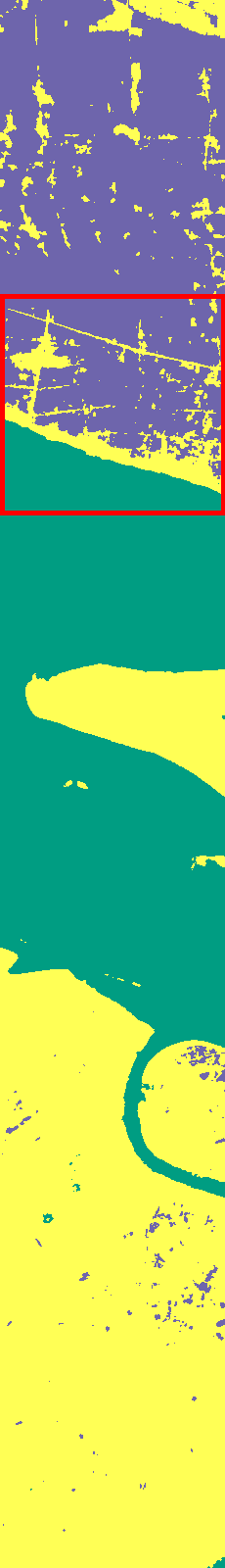}}%
\label{result_shanghang_dst}
\vspace{\resultvspace}
\subfloat[]{\includegraphics[width=0.1\linewidth]{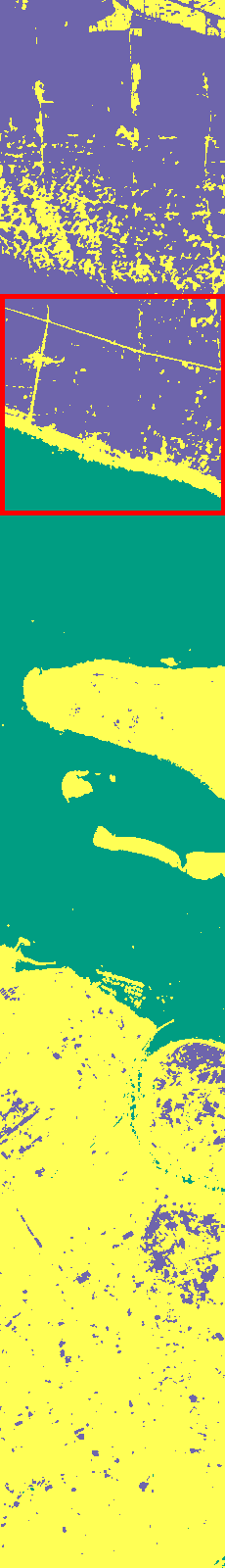}}%
\label{result_shanghang_dsn}
\vspace{\resultvspace}
\subfloat[]{\includegraphics[width=0.1\linewidth]{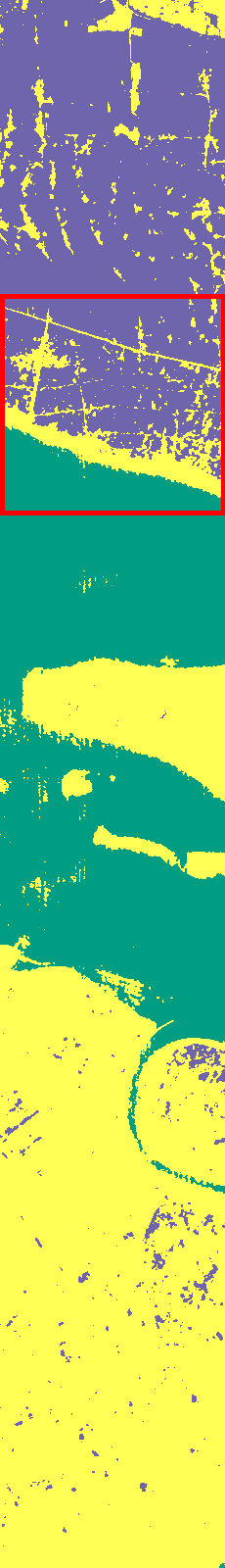}}%
\label{result_shanghang_scluda}
\vspace{\resultvspace}
\subfloat[]{\includegraphics[width=0.1\linewidth]{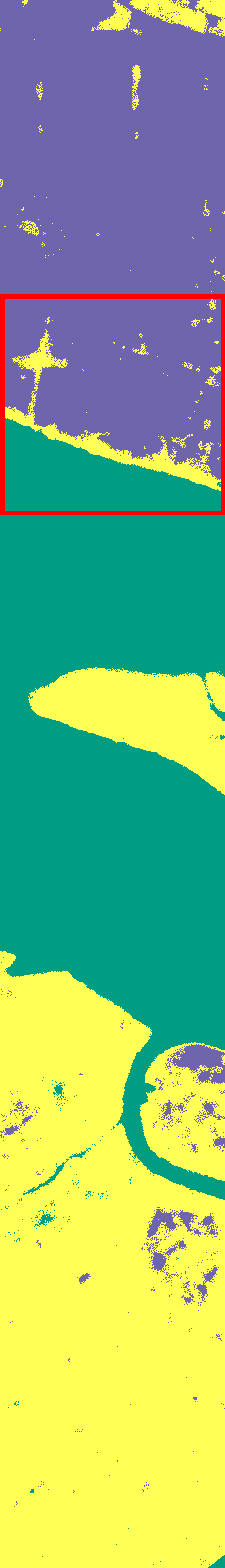}
}
\label{result_shanghang_tstnet}
\vspace{\resultvspace}
\subfloat[]{\includegraphics[width=0.1\linewidth]{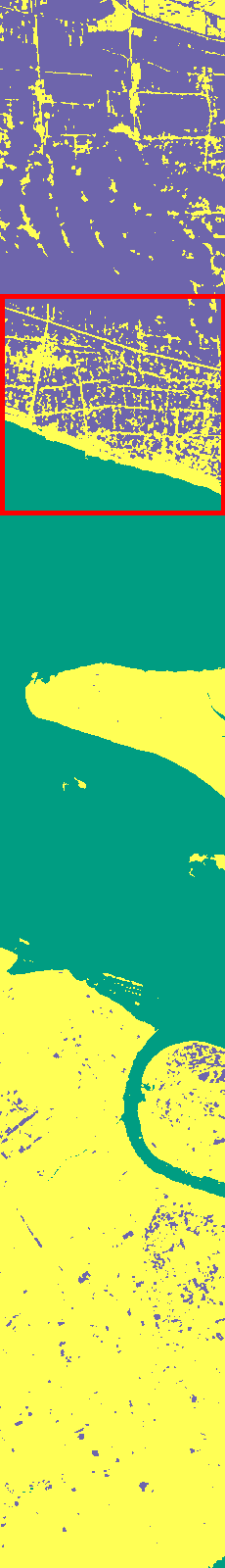}
}
\label{result_shanghang_our}
}

\caption{Classification map for S-H with different methods: (a) SVM; (b) DDC; (c) DAN; (d) JAN; (e) DSAN; (f) DANN; (g) MCD;(h) ST; (i) DSN; (j) SCLUDA; (k) TSTNet and (l) S\textsuperscript{4}DL.}
\label{fig:res_vis_sh}
\end{figure*}

\textbf{S-H.}
As shown in Table.~\ref{tab:res-sh}, compared with other methods, our S\textsuperscript{4}DL exhibits improvements of at least 1.0\% in OA and 2.0\% in Kappa, respectively. 
Especially compared with TSTNet, S\textsuperscript{4}DL maintains a high accuracy with improvements of 10.7\% in OA and 16.2\% in Kappa. When compared with DSN, our method has improved the OA by 1.3\% and the Kappa score by 1.8\%, verifying that our S\textsuperscript{4}DL can alleviate the channel confusion caused by the phenomenon of same objects with different spectra in cross-scene HSIs. 
The visualization in Fig.~\ref{fig:res_vis_sh} shows that S\textsuperscript{4}DL effectively extracts the details and edge information. For example, in the Land/Building of upper half of the image, S\textsuperscript{4}DL retains the integrity of topology structure while reducing domain shifts.

\subsection{Ablation Study}
To verify the effectiveness and contribution of each component to the overall performance by our S\textsuperscript{4}DL, we conducted ablation studies on the three selected datasets.

\begin{table}[ht!]
    \centering
    \caption{Classification performance of each module in ablation experiments}
    \label{tab:res-abl}
    \footnotesize 
    \centering
    
    \begin{tabular}{c|c|c|c|c|c}
    \toprule[1.5pt]
    \textbf{Dataset} & \textbf{GSSD} & \textbf{SSAM} & \textbf{RFE} & \textbf{OA} & \textbf{Kappa} \\
    \midrule[1.0pt]
    \multirow{6}{*}{Houston} 
    &  -       &  -       &  -       & 57.3$\pm$1.5  & 45.6$\pm$1.5 \\
    & $\surd$ &  -       &  -       & 62.6±3.6 & 50.3±3.6 \\
    & $\surd$ & $\surd$ &  -       & 64.5±2.5 & 52.2±2.8 \\
    &     -    &  -       & $\surd$ & 64.7±2.5 & 53.2±2.1 \\
    & $\surd$ &   -      & $\surd$ & 69.5±1.0 & 58.2±1.1 \\
    & $\surd$ & $\surd$ & $\surd$ & \textbf{72.0±2.3} & \textbf{60.4±2.1} \\
    \midrule[1.0pt]
    \multirow{6}{*}{HyRANK} 
    &  -       &   -      & -        & 54.5±3.6 & 49.1±3.6 \\
    & $\surd$ &  -       &   -      & 60.0±2.1 & 55.0±2.3 \\
    & $\surd$ & $\surd$ &  -       & 61.5±1.4 & 56.6±1.4 \\
    &     -    &   -      & $\surd$ & 62.0±1.4 & 57.2±1.5 \\
    & $\surd$ &  -       & $\surd$ & 62.2±1.7 & 57.5±1.8 \\
    & $\surd$ & $\surd$ & $\surd$ & \textbf{65.0±1.9} & \textbf{60.2±2.0} \\
    \midrule[1.0pt]
    \multirow{6}{*}{S-H} 
    &    -     &   -      &   -      & 84.2±3.8 & 77.5±5.1 \\
    & $\surd$ &   -      &  -       & 88.4±5.4 & 83.3±7.2 \\
    & $\surd$ & $\surd$ &  -       & 91.1±2.7 & 87.0±3.8 \\
    &    -     &   -      & $\surd$ & 89.9±3.3 & 85.3±4.5 \\
    & $\surd$ &    -     & $\surd$ & 90.7±1.1 & 86.3±1.5 \\
    & $\surd$ & $\surd$ & $\surd$ & \textbf{92.4±1.2} & \textbf{88.8±1.7} \\
    \bottomrule[1.5pt]
    \end{tabular}
\end{table}

\begin{table}[ht!]
    \centering
    \caption{Classification performance of each banch of GSSD in ablation experiments}
    \label{tab:res-gssd}
    \footnotesize 
    \centering
    
    \begin{tabular}{c|c|c|c|c}
    \toprule[1.5pt]
    \textbf{Dataset} & \textbf{GSSD/\(\mathbf{F}^{di}\)} & \textbf{GSSD/\(\mathbf{F}^{ds}\)} & \textbf{OA} & \textbf{Kappa} \\
    
    \midrule[1.0pt]
    \multirow{4}{*}{Houston} 
    &  -       &  -    & 57.3±1.5  & 45.6±1.5 \\
    & $\surd$  &  -    & 62.1±4.0 & 49.7±4.2 \\
    &   -      &$\surd$& 61.0±2.2 & 48.7±2.5 \\
    & $\surd$  &$\surd$& \textbf{62.6$\pm$3.5}  & \textbf{50.7$\pm$3.2} \\
    
    \midrule[1.0pt]
    \multirow{4}{*}{HyRANK} 
    &  -       &  -    & 54.5±3.6 & 49.1±3.6 \\
    & $\surd$  &  -    & 55.3±3.1 & 50.2±3.2 \\
    &   -      &$\surd$& 55.9±2.8 & 50.8±2.9 \\
    & $\surd$  &$\surd$& \textbf{60.0±2.1}& \textbf{55.0±2.3} \\
    
    \midrule[1.0pt]
    \multirow{4}{*}{S-H} 
    &  -       &  -    & 84.2±3.8 & 77.5±5.1 \\
    & $\surd$  &  -    & 87.9±3.4 & 82.5±4.6\\
    &   -      &$\surd$& 86.2±6.5 & 80.5±8.4 \\
    & $\surd$  &$\surd$& \textbf{88.4±5.4} & \textbf{83.3±7.2} \\
    
    \bottomrule[1.5pt]
    \end{tabular}
\end{table}

DANN, which has the same \(\mathcal{L}_{\textrm{dom}}\) as the proposed S\textsuperscript{4}DL, is selected as the baseline model to verify the effectiveness of each module.
As presented in Table.~\ref{tab:res-abl}, the generalization performance of the baseline model is relatively weak.
When solely using GSSD, in order to eliminate the impact of different mask ratios, we experimented with the fixed mask ratio \(r\) at \(\{0\%, 5\%, 10\%, 15\%, 20\%\}\) and recorded the highest value as the result. The integration of GSSD results in substantial improvements across all the metrics for the three datasets. The OA improved by 5.3\%, 5.5\% and 4.2\%, and the Kappa improved 4.7\%, 5.9\% and 5.8\%, respectively. This demonstrates that models without suitable adaptive strategies struggle with cross-scene HSI classification. 

After confirming the effectiveness of GSSD, we further incorporate SSAM to verify the impact on model performance by dynamically adjusting the disentangling strategy according to the scale of domain shifts between different datasets and different training stages. 
As shown in Table.~\ref{tab:res-abl}, subsequent integration of SSAM leads to varying degrees of improvement on the three selected datasets. The OA scores are improved by 1.9\%, 1.5\% and 2.7\% on three datasets, and the Kappa improved by 1.9\%, 1.6\% and 3.7\%, respectively. The varying degrees of improvement across different datasets demonstrates the capacity of SSAM to dynamically modulate the intensity of alignment in relation to the scale of the domain shift inherent in each dataset and training stage.

Meanwhile, the usage of RFE has improved the baseline by  preserving and embedding domain information in low-level features. 
On this new baseline, adding GGSD and SSAM can further improve the classification performance. Ultimately, the model reaches its peak performance when all the modules are activated. 

Table.~\ref{tab:res-gssd} shows the role of \(\mathbf{F}^{di}\) and \(\mathbf{F}^{ds}\) branches in GSSD and their impact on model performance. It can be seen that the channel decomposition alone in either \(\mathbf{F}^{di}\) or \(\mathbf{F}^{ds}\) improves the model's performance. This confirms the premise that rich domain information exists in the spectral dimension. When \(\mathbf{F}^{di}\) and \(\mathbf{F}^{ds}\) channels are decomposed at the same time, the model performance is best. 
With the addition of GSSD, the model's cross-scene classification ability is significantly enhanced because the joint disentangling strategy can comprehensively decouple different domain-invariant and domain-specific features, ensuring the model's transferability.

\subsection{Feature Visualization}

\begin{figure*}
    \centering
    \subfloat[]{\includegraphics[width=0.24\linewidth]{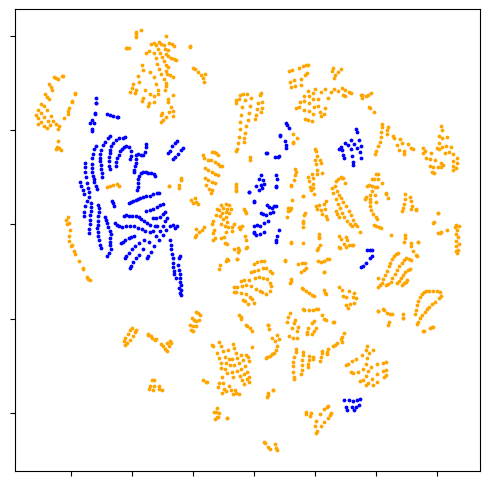}
    \label{fig:tsne-a}}
    \hfill
    \subfloat[]{\includegraphics[width=0.24\linewidth]{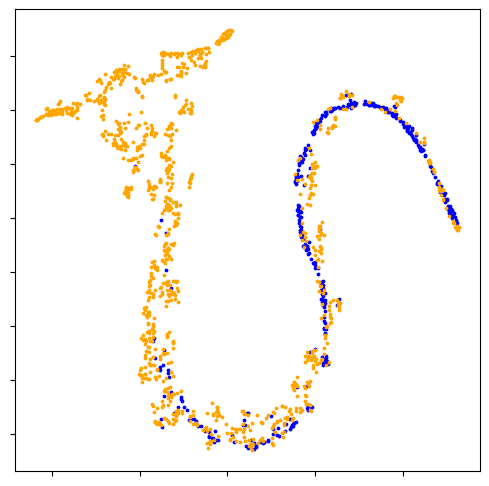}
    \label{fig:tsne-b}}
    \hfill
    \subfloat[]{\includegraphics[width=0.24\linewidth]{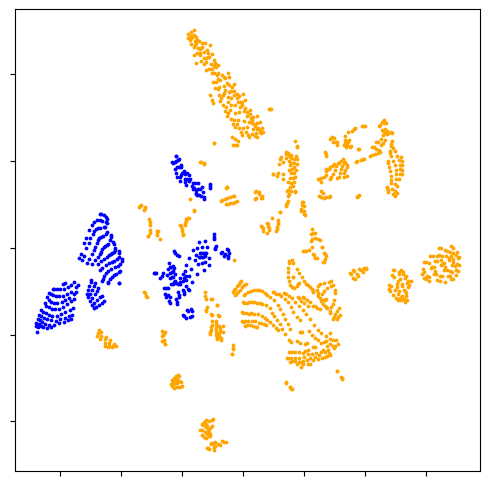}
    \label{fig:tsne-c}}
    \hfill
    \subfloat[]{\includegraphics[width=0.24\linewidth]{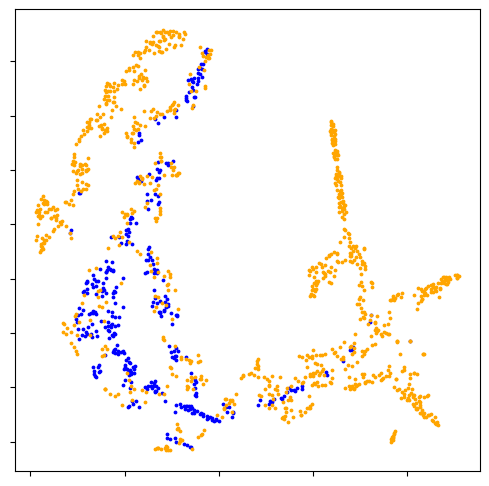}
    \label{fig:tsne-d} }
    
    \subfloat[]{\includegraphics[width=0.24\linewidth]{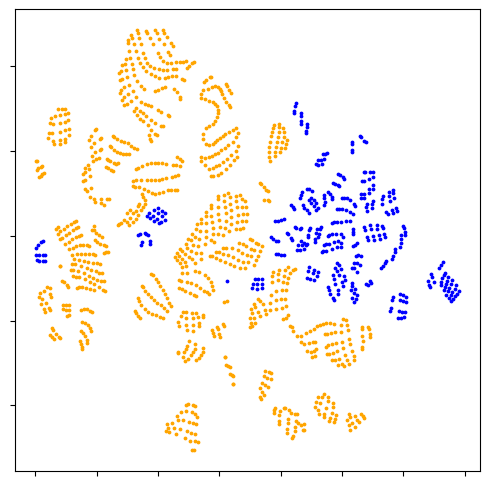}
    \label{fig:tsne-e}}
    \hfill
    \subfloat[]{\includegraphics[width=0.24\linewidth]{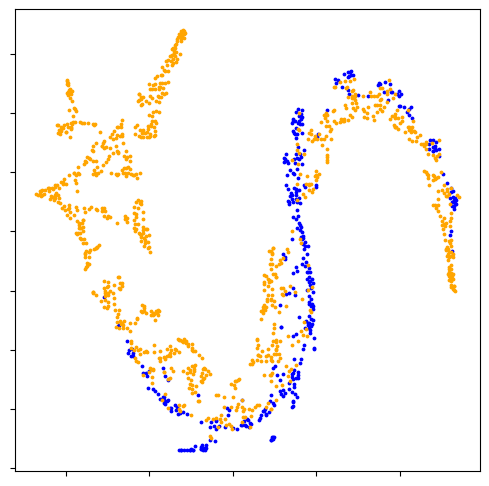}
    \label{fig:tsne-f}
    }
    \hfill
    \subfloat[]{\includegraphics[width=0.24\linewidth]{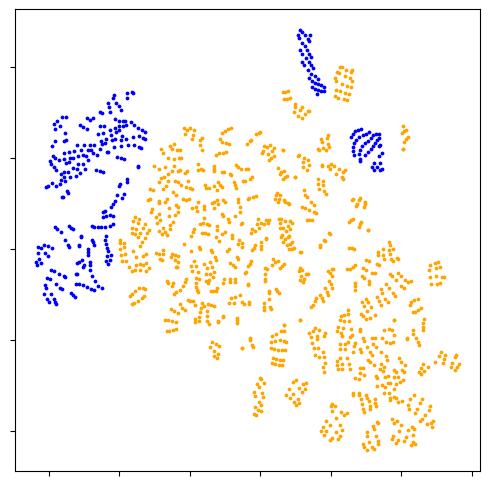}
    \label{fig:tsne-g}}
    \hfill
    \subfloat[]{\includegraphics[width=0.24\linewidth]{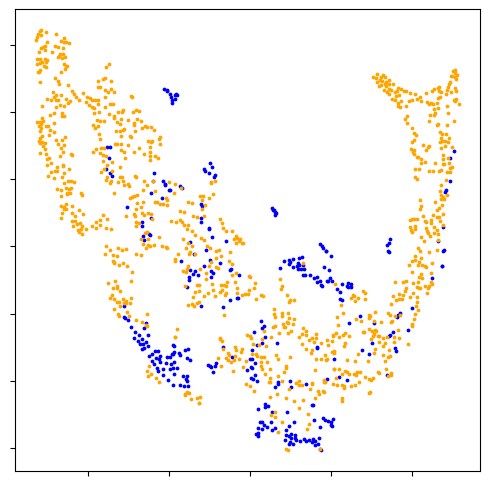}
    \label{fig:tsne-h}}
    \caption{Visualization of alignment performance on Houston dataset. (a) Original samples from Houston 2013 (Class1).  (b) \(\tilde{\mathbf{F}}^{di}\) extracted by S\textsuperscript{4}DL (Class1).  (c) Original samples from Houston 2013 (Class2). (d) \(\tilde{\mathbf{F}}^{di}\) extracted by S\textsuperscript{4}DL (Class2). (e) Original samples from Houston 2013 (Class3). (f) \(\tilde{\mathbf{F}}^{di}\) extracted by S\textsuperscript{4}DL (Class3). (g) Original samples from Houston 2013 (Class5). (h) \(\tilde{\mathbf{F}}^{di}\) extracted by S\textsuperscript{4}DL (Class5). }
    \label{fig:tsne}
\end{figure*}

To further assess the alignment performance, we use t-SNE to reduce dimensionality and visualize the distribution of the input data and the domain-invariant feature \(\tilde{\mathbf{F}}^{di}\) extracted by S\textsuperscript{4}DL on the Houston dataset, as shown in Fig.~\ref{fig:tsne}. Fig.~\ref{fig:tsne-a},~\ref{fig:tsne-c},~\ref{fig:tsne-e} and~\ref{fig:tsne-g} depict the distributions for three different classes, while Fig.~\ref{fig:tsne-b},~\ref{fig:tsne-d},~\ref{fig:tsne-f} and~\ref{fig:tsne-h} display the distributions of \(\tilde{\mathbf{F}}^{di}\). The distribution of source domain data or \(\tilde{\mathbf{F}}^{di}\) is represented in blue, and the distribution of target domain data or \(\tilde{\mathbf{F}}^{di}\) is represented in orange. All the data is mapped to 2D by t-SNE method.

It can be clearly observed that in the original samples, there is a significant domain shift between the distributions of the source domain and the target domain.  Interestingly, following feature extraction by S\textsuperscript{4}DL,  there is some overlap in  \(\tilde{\mathbf{F}}^{di}\) from different domains, and the  distribution of identical categories in \(\tilde{\mathbf{F}}^{di}\) tends to be consistent.  This denotes that the features of both the source and target domain in \(\tilde{\mathbf{F}}^{di}\) align to the same feature space, effectively  alleviating the domain shift.

\begin{figure*}
    \centering
    \subfloat[]{\includegraphics[width=0.32\linewidth]{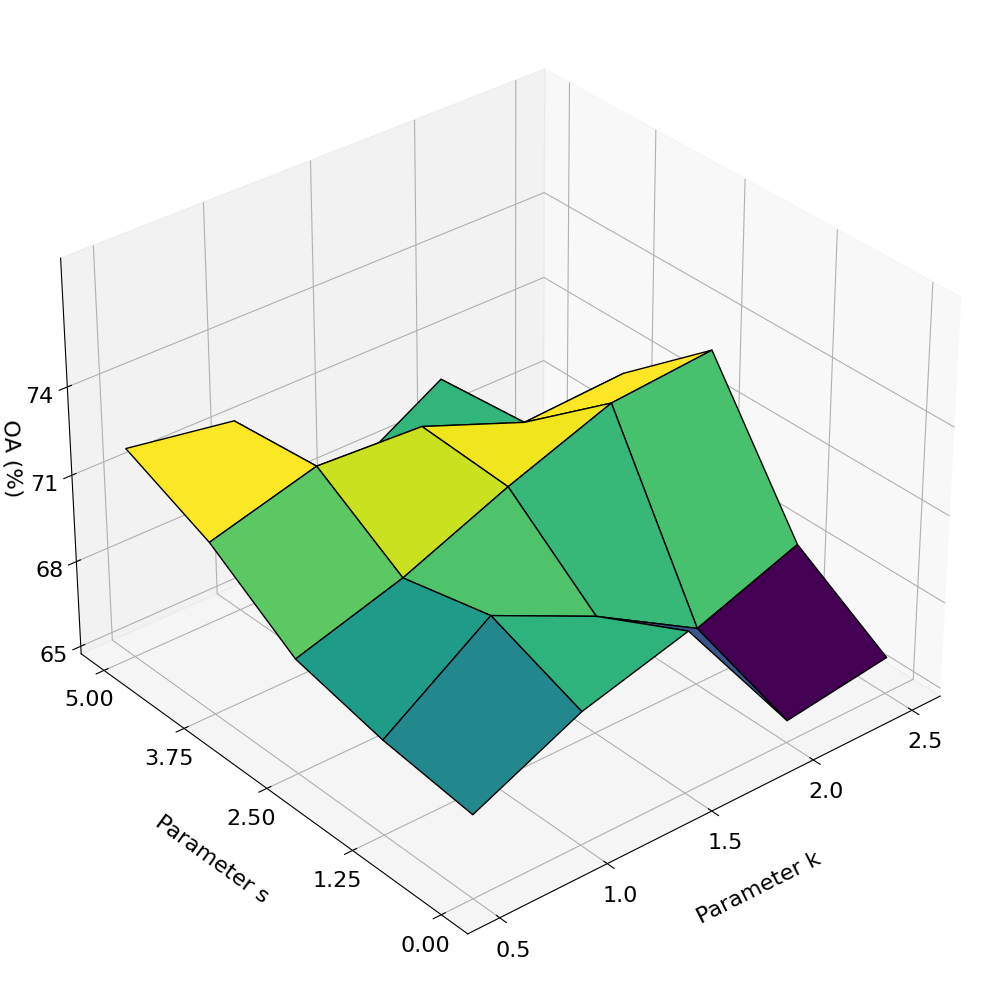}}
    \hfill
    \subfloat[]{\includegraphics[width=0.32\linewidth]{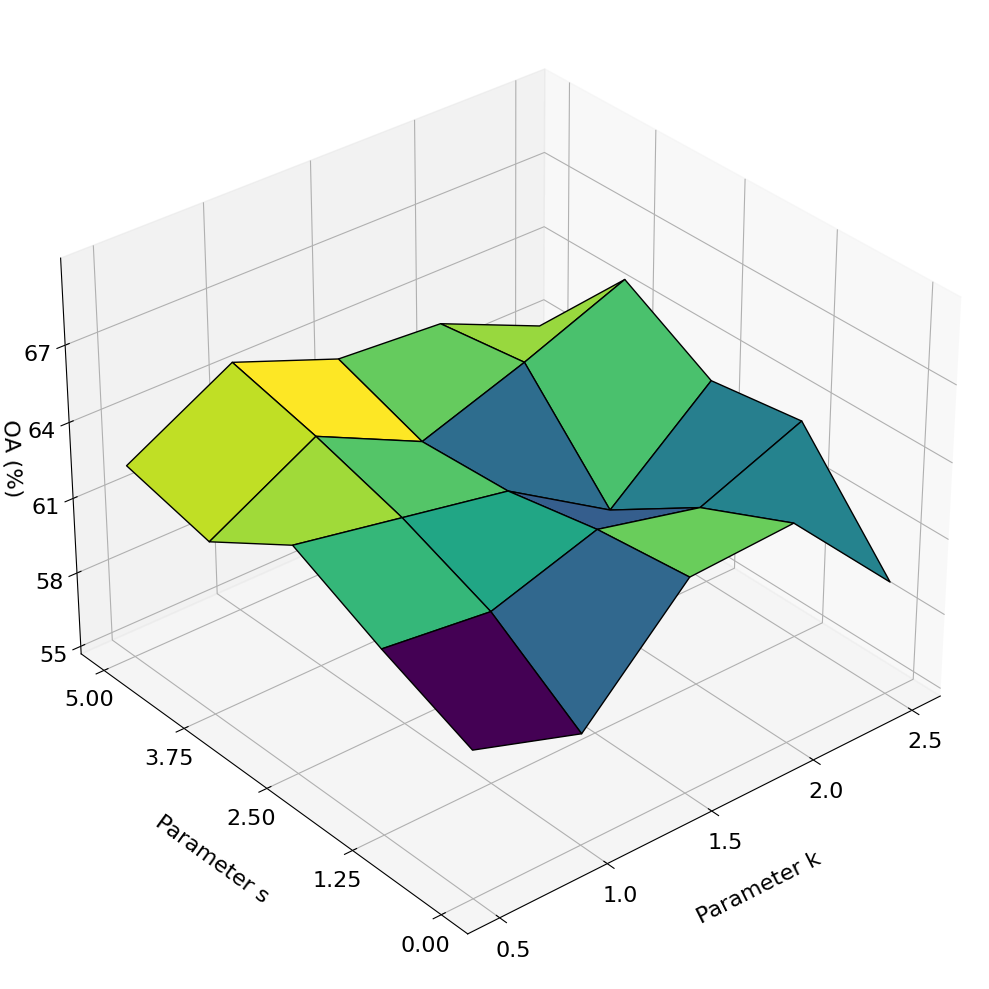}}
    \hfill
    \subfloat[]{\includegraphics[width=0.32\linewidth]{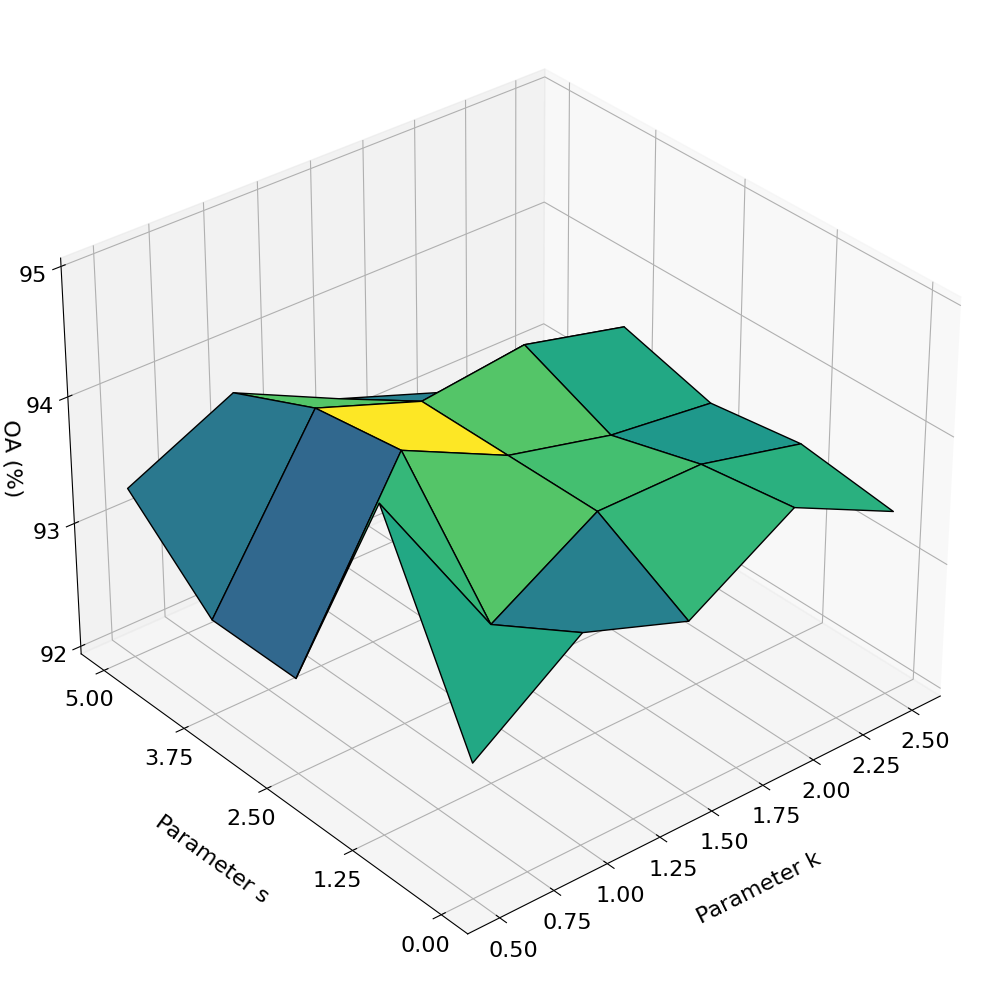}}
    \hfill
    \caption{Parameter tuning of the offsets \(k\) of the slopes \(s\) of Sigmoid function. (a) Houston dataset. (b) HyRANK dataset. (c) S-H dataset.}
    \label{fig:3d}
\end{figure*}

\subsection{Parameter Tuning}
\label{sec: pt}
In S\textsuperscript{4}DL, the slopes \(k\) and the offsets \(s\) determine the initial value and speed of mask ratio \(r_e\) updated in SSAM. Therefore, this determines the intensity of channel disentangling, thus the model is rather sensitive to the choice of \(k\) and \(s\). To analyse parameter sensitivity of S\textsuperscript{4}DL on three datasets, the grid search is conducted for different parameters. The search range for \(k\) is {0.5, 1, 1.5, 2, 2.5}, and the search range for \(s\) is {0, 1.25, 2.5, 3.75, 5}. Fig.~\ref{fig:3d} shows the change trend of classification results of S\textsuperscript{4}DL with different parameters on three datasets. 

It can be seen that when \(s\) is fixed and \(k\) is in the interval [0.5, 1.5], OA rises as \(k\) increases. This is because as \(k\) gradually increases, the intensity of model disentangling can be updated more quickly. While  \(k\) is in the interval [1.5, 2.5], OA decreases with the increase of \(k\). This is because as \(k\) becomes too large, the large fluctuation of the disentangling intensity causes the training process to be unstable, thereby deteriorating the model performance.

\begin{table*}
    \caption{Number of parameters (M, million) in different methods}
    \label{tab:eff}
    \centering
    \resizebox{\textwidth}{!}{
    \begin{tabular}{c|c|c|c|c|c|c|c|c|c|c|c|c} 
    \toprule[1.5pt]
    \multicolumn{2}{c|}{Method} & DDC~\cite{tzeng2014deep} & DAN~\cite{long2015learning} & JAN~\cite{long2017deep} & DSAN~\cite{zhu2020deep} & DANN~\cite{ganin2015unsupervised} & MCD~\cite{saito2018maximum} & ST~\cite{zou2018unsupervised} & DSN~\cite{bousmalis2016domain} & SCLUDA~\cite{li2023supervised} & TSTNet~\cite{zhang2021topological}  & S\textsuperscript{4}DL \\
    \midrule[1.0pt]
    \multicolumn{2}{c|}{\#params(M)} & 11.48 & 11.48 & 11.48 & 11.48 & 11.63 & 11.63 & 11.79 & 35.85 & 2.05 & 7.83 & 1.06 \\
    \midrule[1.0pt]
    \multirow{3}{*}{\#FLOPs(M)} 
    & Houston & 19.51 & 19.51 & 19.51 & 19.51 & 19.67 & 19.67 & 19.82 & 48.89 & 37.50 & 8.49 & 3.00 \\
    & HyRANK & 33.96 & 33.96 & 33.96 & 33.96 & 34.12 & 34.12 & 34.28 & 86.72 & 140.10 & 12.18 & 8.31 \\
    & S-H    & 36.44 & 36.44 & 36.44 & 36.44 & 36.60 & 36.60 & 36.76 & 93.22 & 157.73 & 12.81 & 9.11 \\
    \bottomrule[1.5pt]
    \end{tabular}
    }
\end{table*}

Correspondingly, when \(k\) is fixed within the interval [0, 2.5], the model performance improves with the increase of \(s\). 
This is because when \(s\) increases, the value of the mask ratio correspondingly decreases, which ensures the learning of discriminative features. Within  \(k\) is the interval [2.5, 5], the model performance decreases with the rise of \(s\). This is because when the mask ratio is too low, the model cannot completely disentangle the domain-invariant channel and the domain-specific channel, thereby reducing the transfer performance.

\subsection{Model Efficiency}
The number of parameters and FLOPs of different methods are listed in Table.~\ref{tab:eff} to compare the computational complexity of S\textsuperscript{4}DL with other methods.  
ResNet18~\cite{he2016deep} is chosen as the backbone. The experimental environment and parameter settings, such as patch size, are consistent with Section.~\ref{sec:exp}.~\ref{sec:exp-setting}. The number of parameters and FLOPs of DDC, DAN, JAN, and DSAN are the same because these methods have the same model structure and only differ in loss function. DANN and MCD are slightly increased in terms of parameters and FLOPs due to the addition of a domain discriminator and a classifier, respectively. ST has slight increases in the quantities as well due to their three classifiers. DSN has the most parameters among all the methods because it includes three encoders and one decoder. SCLUDA has the largest FLOPs due to its 3D convolutions and data augmentation strategy. Due to its simple CNN and GCN structure, TSTNet has low parameters and FLOPs. 
Due to the the efficient binary kernel in GSSD and the well-designed, simple RFE, our S\textsuperscript{4}DL has the lowest parameters and the least FLOPs on all the datasets.

\section{Conclusion}
In this paper, a novel and efficient shift-sensitive joint disentangling learning framework S\textsuperscript{4}DL is proposed for cross-scene HSI classification. For the quantitation and separation of domain invariant and domain-specific information in spatial-spectral dimension, S\textsuperscript{4}DL constructs the GGSD. The dynamic feature decomposition allows the model to extract more comprehensive domain invariant features for cross-domain classification. For adaptation to the various scales of domain shift of different datasets and different training stages, S\textsuperscript{4}DL designed the SSAM that adjusts the disentangling strategy in real-time, improving the model's generalization on different datasets. For preservation of domain information in low-order features, S\textsuperscript{4}DL ensembles the RFE to retain and embed low-level features while extracting high-level features. Experimental results on three commonly used cross-scene HSI datasets demonstrate that the proposed S\textsuperscript{4}DL achieves better transfer performance than many other state-of-the-art methods.

\ifCLASSOPTIONcaptionsoff
  \newpage
\fi


%
\bibliographystyle{IEEEtran}
\bibliography{IEEEabrv,main}

\end{document}